\documentclass[preprint,12pt]{elsarticle}
\usepackage{bm}
\newcommand{\KL}{\textup{KL}}
\newcommand{\ELBO}{\textup{ELBO}}
\newcommand{\MMD}{\textup{MMD}}
\newcommand{\KID}{\textup{KID}}
\newcommand{\IB}{\textup{IB}}
\newcommand{\BF}{\textup{BF}}
\newcommand{\LF}{\textup{LF}}
\newcommand{\HF}{\textup{HF}}
\newcommand{\I}{\mathbb{I}}
\newcommand{\Err}{\mathcal{E}}

\usepackage{amsfonts}
\usepackage{amsthm,physics}
\usepackage{amsmath}
\usepackage{amscd}
\usepackage{t1enc}
\usepackage[mathscr]{eucal}
\usepackage{indentfirst}
\usepackage{graphicx}
\usepackage{graphics}
\usepackage{pict2e}
\usepackage{epic}
\numberwithin{equation}{section}
\usepackage{epstopdf}
\usepackage{amsmath,amsthm,verbatim,amssymb,amsfonts,amscd,graphicx}
\usepackage{mathtools}
\usepackage[dvipsnames]{xcolor}
\usepackage{bm}
\usepackage{soul}

\usepackage[dvipsnames]{xcolor}
\usepackage[margin=3cm]{caption}
\usepackage{amsmath}
\usepackage{amsthm}
\usepackage{amssymb}
\usepackage{amscd}
\usepackage{mathtools}
\usepackage{hyperref}
\usepackage{subfigure}
\usepackage{stmaryrd}
\usepackage{booktabs}
\usepackage{mdframed}
\usepackage{upgreek}
\usepackage{mathtools} \mathtoolsset{showonlyrefs=true} \MakeRobust{\eqref}
\usepackage{listings}
\usepackage{color}
\usepackage{fullpage}
\usepackage{stackrel}
\usepackage{textcomp}
\usepackage{enumerate}
\usepackage{accents}
\usepackage{graphicx}
\usepackage{epstopdf}
\usepackage{placeins}
\usepackage{bbm}
\usepackage{bm}
\usepackage{accents}
\usepackage[mathscr]{eucal}
\usepackage[linesnumbered,boxed,titlenumbered,ruled,nofillcomment]{algorithm2e}
\usepackage[T1]{fontenc}
\usepackage{lmodern}
\usepackage{scalerel}
\usepackage{wrapfig}
\usepackage{algorithmic}
\usepackage{float}

\definecolor{lbcolor}{rgb}{0.9,0.9,0.9}
\theoremstyle{plain}
\newtheorem{theorem}{Theorem} \numberwithin{theorem}{section}

\newtheorem{prop}[theorem]{Proposition}

\theoremstyle{definition}

\newtheorem{definition}[theorem]{Definition}

\definecolor{mydarkblue}{rgb}{0,0.08,0.45}
\hypersetup{ %
	colorlinks=false,
	linkcolor=mydarkblue,
	citecolor=mydarkblue,
	filecolor=mydarkblue,
	urlcolor=mydarkblue}

\newcommand{\Fc}[0]{\mathcal{F}}
\newcommand{\Gc}{\mathcal{G}}
\newcommand{\Hc}{\mathcal{H}}

\definecolor{mypink1}{rgb}{0.858, 0.188, 0.478}
\definecolor{mypink2}{RGB}{219, 48, 122}
\definecolor{mypink3}{cmyk}{0, 0.7808, 0.4429, 0.1412}
\definecolor{mygray}{gray}{0.6}

\newlength{\leftstackrelawd}
\newlength{\leftstackrelbwd}
\def\leftstackrel#1#2{\settowidth{\leftstackrelawd}%
{${{}^{#1}}$}\settowidth{\leftstackrelbwd}{$#2$}%
\addtolength{\leftstackrelawd}{-\leftstackrelbwd}%
\leavevmode\ifthenelse{\lengthtest{\leftstackrelawd>0pt}}%
{\kern-.5\leftstackrelawd}{}\mathrel{\mathop{#2}\limits^{#1}}}

\def\R{{\mathbb R}}

\def\E{{\mathbb E}}
\def\R{{\mathbb R}}

\def\P{{\mathbb P}}

\def\u{{\mathbf{u}}}

\newcommand{\bs}[1]{\boldsymbol{#1}}

\DeclareMathOperator*{\argmax}{argmax} 

\journal{TBA}

\begin{document}

\begin{frontmatter}

\title{Bi-fidelity Variational Auto-encoder for Uncertainty Quantification}

\author[1]{Nuojin Cheng}
\author[2]{Osman Asif Malik}
\author[3]{Subhayan De}
\author[1]{Stephen Becker}
\author[4]{Alireza Doostan}

\affiliation[1]{organization={Department of Applied Mathematics, University of Colorado, Boulder}
}

\affiliation[2]{organization={Applied Mathematics \& Computational Research Division, Lawrence Berkeley National Laboratory}}

\affiliation[3]{organization={Department of Mechanical Engineering, Northern Arizona University}}

\affiliation[4]{organization={Ann and H.J. Smead Department of Aerospace Engineering Sciences, University of Colorado, Boulder}}

\begin{abstract}
    
Quantifying the uncertainty of quantities of interest (QoIs) from physical systems is a primary objective in model validation. However, achieving this goal entails balancing the need for computational efficiency with the requirement for numerical accuracy.
To address this trade-off, we propose a novel bi-fidelity formulation of variational auto-encoders (BF-VAE) designed to estimate the uncertainty associated with a QoI from low-fidelity (LF) and high-fidelity (HF) samples of the QoI. 
This model allows for the approximation of the statistics of the HF QoI by leveraging information derived from its LF counterpart. Specifically, we design a bi-fidelity auto-regressive model in the latent space that is integrated within the VAE's probabilistic encoder-decoder structure. An effective algorithm is proposed to maximize the variational lower bound of the HF log-likelihood in the presence of limited HF data, resulting in the synthesis of HF realizations with a reduced computational cost. Additionally, we introduce the concept of the bi-fidelity information bottleneck (BF-IB) to provide an information-theoretic interpretation of the proposed BF-VAE model. Our numerical results demonstrate that BF-VAE leads to considerably improved accuracy, as compared to a VAE trained using only HF data, when limited HF data is available.
\end{abstract}

\begin{keyword}
uncertainty quantification \sep variational auto-encoder \sep multi-fidelity \sep generative modeling \sep transfer learning
\end{keyword}

\end{frontmatter}


%
%
\section{Introduction}

Uncertainty pervades numerous engineering applications due to various factors, such as material properties, operating environments, and boundary conditions, which impact the prediction of a performance metric or quantity of interest (QoI), denoted as $\bm{x}\in\mathbb{R}^D$, following an unknown probability density function (pdf) $p(\bm x)$. The quantification of uncertainty in $\bm{x}$ through the estimation of its moments or distribution has been an active area of research within the field of uncertainty quantification (UQ). One approach to accomplish this involves collecting independent and identically distributed (i.i.d.) realizations of $\bm{x}$ to estimate its empirical properties.
However, when the realizations of $\bm x$ are obtained through the solution of computationally intensive models, generating a large enough set of realizations to ensure statistical convergence becomes infeasible. To address this challenge, a surrogate model of the forward map between uncertain inputs $\bm\xi\in\R^M$ and $\bm x$ can be constructed. This approach has been demonstrated through a range of techniques, including polynomial chaos expansion \cite{ghanem2003stochastic,peng14weighted,hampton2015compressive,shustin2022pcenet}, Gaussian process regression \cite{williams2006gaussian,bilionis2012multi}, and deep neural networks \cite{tripathy2018deep,zhu2019physics,padmanabha2021solving,shustin2022pcenet}. Once established, the surrogate model can be employed, often at a negligible cost, to generate realizations of the QoI and estimate its statistics. However, it should be noted that the complexity of constructing these surrogate models often increases rapidly with the number of uncertain variables, $M$, a phenomenon referred to as the curse of dimensionality.

To mitigate the problem caused by high-dimensional uncertainty, one remedy is to build a reduced-order model (ROM),  \cite{hesthaven2016certified}, where the solution to the governing equations is approximated in a basis of size possibly independent of $M$. One widely adopted technique for identifying such a reduced basis is proper orthogonal decomposition (POD), often also referred to as principle component analysis (PCA) or Karhunen–Loéve expansion \cite{chatterjee2000introduction}. 
POD is commonly employed on a collection of forward problem simulations, known as snapshots, to determine the optimal subspace via the solution of a singular value decomposition problem.

The utility of ROMs has been extensively investigated for problems that exhibit a small Kolmogorov $n$-width \cite{pinkus2012n}, e.g., diffusion-dominated flows. However, for advection-dominated problems where solutions do not align closely with any linear subspace, conventional ROMs may yield inaccurate approximations. 
This has resulted in the development of non-linear (manifold-based) ROM formulations, including kernel principal component analysis \cite{zhou2020kernel,razi2021kernel}, tangent space alignment \cite{zhang2004principal}, and auto-encoders (AEs) \cite{lee2020model,maulik2021reduced,nikolopoulos2022non}. 
Among these manifold-based ROMs, AEs have gained significant attention due to their expressive neural-network-based encoder-decoder structure, enabling them to capture the underlying patterns of the input data by learning a latent representation. 
The latent variable, denoted by $\bm z$, is of much lower dimension than the input data and is learned through a non-linear encoder function. 
The decoder function, which typically has a structure mirroring the encoder, takes $\bm z$ and maps it back to the original data space.

While AE-based UQ models \cite{lee2020model,maulik2021reduced,nikolopoulos2022non} have demonstrated success, they are intrinsically deterministic as they do not automatically produce new samples of $p(\bm x)$ and, therefore, are not generative. Furthermore, as shown in \cite{steck2020autoencoders}, the lack of regularization in the fully-connected AE architecture can lead to overfitting,  hindering the discovery of meaningful latent representations.  To address these limitations, several probabilistic frameworks have been proposed to regularize the problem and, more importantly, enable uncertainty estimation. These include Bayesian convolutional AE \cite{zhu2018bayesian} designed specifically for flow-based problems and the auto-regressive encoder-decoder model for turbulent flows \cite{geneva2020modeling}. 

In this study, we consider the use of variational autoencoders (VAEs) and present a novel training strategy aimed at reducing the training cost in terms of the number of high-fidelity realizations required. VAEs belong to a class of machine learning models that seek to approximate the unknown underlying distribution $p(\bm x)$ from which the QoI is derived and generate new realizations from it. 
Deep generative models, including VAEs \cite{kingma14auto,rezende2014stochastic}, generative adversarial networks (GANs) \cite{goodfellow14gan}, normalizing flows \cite{rezende15normalizingflow}, and diffusion models \cite{ho2020denoising,song2020score}, have achieved significant success in various applications in computer vision and natural language processing \cite{wang2021generative,petrovich2021action,rajeswar2017adversarial}.
VAEs, in particular, offer a well-suited solution for UQ problems owing to their ability to encode a low-dimensional representation of the QoI, regularized via the probabilistic formulation, and generate new realizations of the QoI. Further details on the VAE methodology can be found in Section~\ref{ssec:vae}.


Despite their benefits, training deep generative models, such as VAEs, typically requires access to a substantial amount of high-fidelity (HF) data, which may be difficult to obtain in large-scale scientific applications. To address this issue, we propose a bi-fidelity VAE (BF-VAE) training approach that leverages a relatively small set of HF realizations of the QoI, $\bm x^H\in\R^D$, along with a larger set of cheaper, possibly less accurate, low-fidelity (LF) realizations of the QoI, $\bm x^L\in\R^D$. By primarily utilizing the LF realizations and establishing low-dimensional mappings between LF and HF variables, we construct a VAE model that captures the underlying distribution of the HF QoI. These low-dimensional mappings are key in reducing the number of HF realizations required for training. 

\begin{figure}[ht]
	\centering
	\includegraphics[width = 0.6\textwidth]{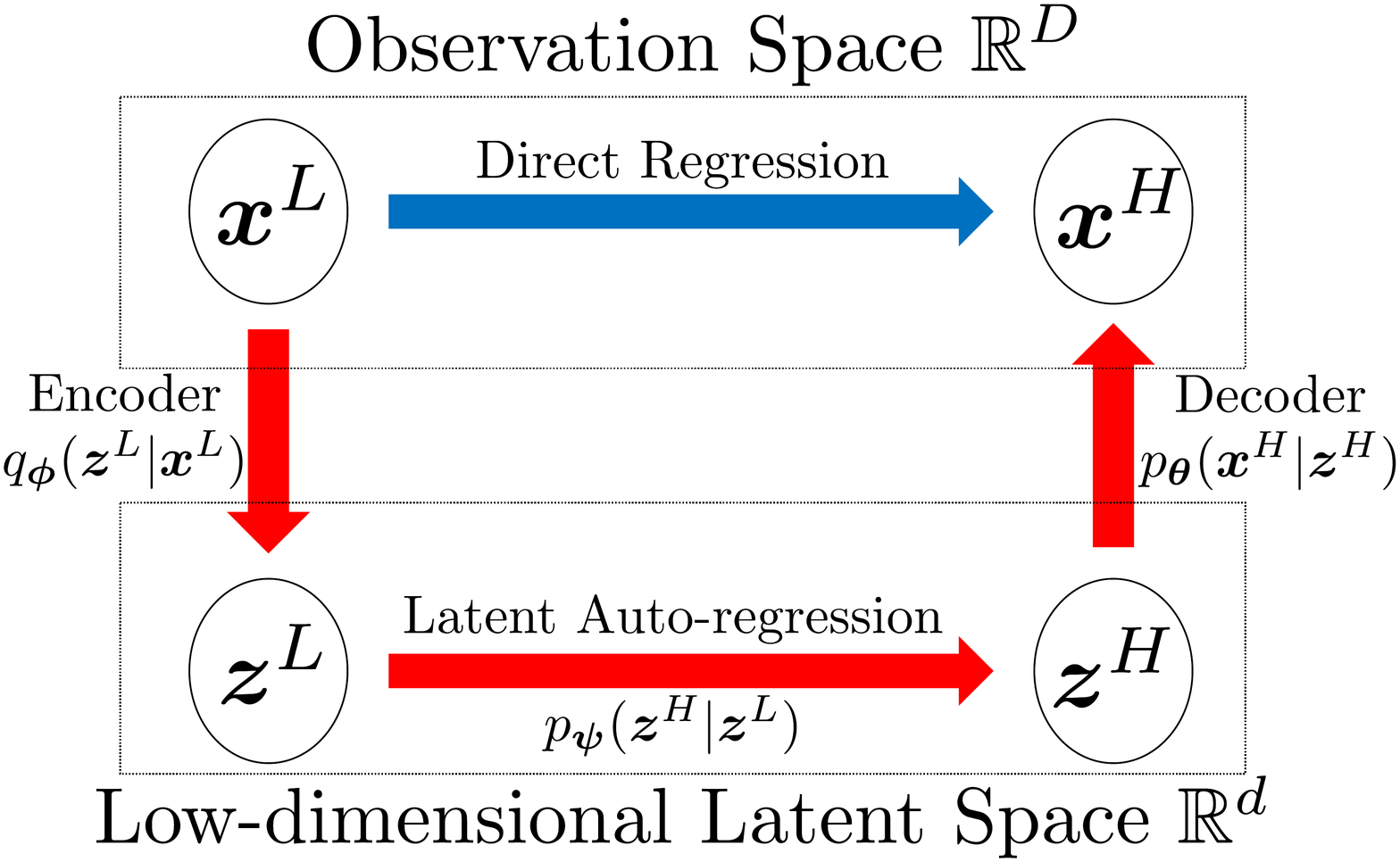}
	\caption{Instead of conducting bi-fidelity regression directly in high-dimensional observation space (blue path), we introduce an approach via low-dimensional latent space (red path).}
	\label{fig:bfvae-graph}
\end{figure}

In more detail, we train a VAE, with the same architecture and activation functions as in the intended HF model, but using LF data. Let $\bm z^L$ and $q_{\bm\phi}(\bm z^L\vert\bm x^L)$ denote the latent variable and encoder of this model. BF-VAE adapts this VAE using HF data in two ways. Firstly, we assume an auto-regressive model with pdf $p_{\bm\psi}(\bm z^H\vert\bm z^L)$, parameterized by $\bm\psi$, to set the latent variable of the HF model, $\bm z^H$. As depicted in Figure~\ref{fig:bfvae-graph}, such a regression is performed in the $d$-dimensional latent space, instead of the observation space of dimension $D\gg d$. Secondly, the subset of the decoder parameters $\bm\theta$ corresponding to the last layer of the decoder $p_{\bm\theta}(\bm x^H\vert\bm z^H)$ are updated (with warm start) to adjust the map between the latent and observation space of the HF data. We note that the latter update also involves a relatively small set  of parameters.  

To summarize, the core contributions of this work are:

\begin{itemize}
    \item We introduce a novel approach --- dubbed BF-VAE --- for training a VAE model, utilizing primarily LF data and a small set of HF data. While trained using both low- and high-fidelity data, BF-VAE aims at approximating the pdf of the HF QoI $\bm x^H$. This, in turn, enables the generation of approximate samples of $\bm x^H$.
    \item The BF-VAE model is theoretically 
    motivated as the maximizer of a training objective criterion we call BF evidence lower bound (BF-ELBO), an extension of the original ELBO formulation introduced in \cite{kingma14auto}.
    We then extend the information bottleneck theory of Tishby et al.\ \cite{tishby2000information} to
    formulate the bi-fidelity information bottleneck (BF-IB) theory and provide an interpretation of BF-ELBO from an information-theoretic perspective.
    \item We conduct an empirical evaluation of the BF-VAE model through three numerical experiments, comparing its performance with a VAE traind only on HF data. The numerical results indicate that BF-VAE improves the accuracy of learned HF QoI pdf when the number of HF data is small.
\end{itemize}

The rest of the paper is structured as follows.
In Section~\ref{sec:background}, we provide an overview of the VAE and linear auto-regressive methods for bi-fidelity regression.
Section~\ref{sec:methodology} elaborates on the proposed BF-VAE model, along with a theoretical interpretation.
The implementation details of the BF-VAE model, including prior density selection and hyperparameter tuning, are presented in Section~\ref{sec:implementation}.
Section~\ref{sec:experiments} showcases three numerical examples that demonstrate the performance of BF-VAE.
Our conclusions are summarized in Section~\ref{sec:conclusion}.
The proof of our main statements and an introduction to an evaluation metric used to assess the quality of data generated from the VAE models are presented in the appendix.

For consistency with other VAE-related papers, we do not differentiate random vectors and their realizations in this work.
Additionally, we simplify density functions by omitting their subscripts. 
For example, we use $p(\bm x\vert\bm y)$ instead of $p_{\bm x\vert\bm y}(\bm x\vert\bm y)$. As such, the densities $p(\bm x^H)$ and $p(\bm x^L)$ may not be the same. 

\section{Motivation and Background}
\label{sec:background}
Forward UQ is concerned with quantifying the uncertainty of QoIs from a physical system due to intrinsic variations or limited knowledge about model inputs (or structure). Within a BF setting, an HF model generates $\bm x^H\in\R^D$ with pdf $p(\bm x^H)$ and an LF model generates an approximation to $\bm x^H$ denoted by $\bm x^L\in\R^D$ and following pdf $p(\bm x^L)$. One goal of forward UQ is to estimate $p(\bm x^H)$.
With a random input vector $\bm\xi\in\R^M$ and its pdf $p(\bm\xi)$, the widely-adopted surrogate modeling approaches seek to approximate $p(\bm x^H\vert\bm\xi)$ and estimate $p(\bm x^H)$ as 
\begin{align}
    p(\bm x^H) &= \int_{\R^M} p(\bm x^H\vert\bm\xi)p(\bm\xi)d\bm\xi.
\end{align}
This formulation, however, suffers from two major issues: the complexity of building $p(\bm x^H\vert\bm \xi)$ when the dimension of $\bm\xi$ is high and the expensive cost of collecting HF QoI realizations $\bm x^{H}$ for estimating $p(\bm x^H\vert\bm \xi)$. 
The Bayesian multi-fidelity approach of \cite{koutsourelakis2009accurate, nitzler2022generalized}, summarized next, provides a solution to tackle these aforementioned issues. 
To alleviate the first issue, bi-fidelity approaches usually introduce an LF pdf $p(\bm x^L)$ with cheaper sampling cost, approximate $p(\bm x^H\vert \bm x^L)$ instead of $p(\bm x^H\vert\bm\xi)$ due to a closer relation between $\bm x^H$ and $\bm x^L$, and marginalize the random input $\bm\xi$ to mitigate the effect of its high-dimensionality,
\begin{equation}
\begin{aligned}
    p(\bm x^H) &= \int_{\R^D}\int_{\R^M} p(\bm x^H,\bm x^L,\bm\xi)d\bm\xi d\bm x^L && \textup{introduce LF model}\\
               &= \int_{\R^D}\int_{\R^M} p(\bm x^H,\bm\xi\,\vert \bm x^L)p(\bm x^L)d\bm\xi d\bm x^L && \textup{condition on LF model}\\
               &= \int_{\R^D} p(\bm x^H\vert \bm x^L)p(\bm x^L)d\bm x^L. && \textup{marginalization}
\end{aligned}
\end{equation}
For the second issue, ROMs introduce a low-dimensional latent variable $\bm z\in\R^d$ with $d\ll D$ to establish a connection between the LF and HF models via $p(\bm x^H\vert\bm x^L)$,
\begin{align}
    p(\bm x^H) &= \int_{\R^D} p(\bm x^H\vert \bm x^L)p(\bm x^L)d\bm x^L\\
    &= \int_{\R^D}\int_{\R^d} p(\bm x^H\vert\bm z, \bm x^L)p(\bm z\vert \bm x^L)p(\bm x^L)d\bm z d\bm x^L.
\end{align}
The latent variable $\bm z$ determines a low-dimensional representation of $\bm x$ that captures the relationship between the LF and HF QoIs, possibly with considerably less HF data for training~ \cite{koutsourelakis2009accurate, nitzler2022generalized}. 

In this work, we assume that the Markov property $p(\bm x^H\vert \bm z,\bm x^L)=p(\bm x^H\vert\bm z)$ holds, which leads to an AE model for $p(\bm x^H)$, given by
\begin{align}
    p(\bm x^H) = \int_{\R^D}\int_{\R^d} \underbrace{p(\bm x^H\vert\bm z)}_{\textup{decoder}}\underbrace{p(\bm z\vert \bm x^L)}_{\textup{encoder}}p(\bm x^L)d\bm z d\bm x^L.
\end{align}
Once the AE model is built, the HF QoI can be estimated following
\begin{align}
    p(\bm x^H) &=  \int_{\R^d} p(\bm x^H\vert\bm z)p(\bm z)d\bm z.
\end{align}
%

This work aims at generating new (approximate) samples from $p(\bm x^H)$ -- rather than deriving an explicit representation $p(\bm x^H)$ -- using which we estimate statistical properties of $\bm x^H$, e.g., $\mathbb{E}[\bm x^H]$ and $\textup{Cov}[\bm x^H]$. This involves three key components, namely, an encoder $p(\bm z\vert\bm x^L)$, the latent variable $p(\bm z)$, and a decoder $p(\bm x^H\vert\bm z)$. In the remaining of this section, we discuss two main ingredients to construct these components. 
In Section~\ref{ssec:vae}, we introduce a VAE approach to building the encoder and decoder in a Bayesian setting. 
In Section~\ref{ssec:auto-regressive}, we discuss an option to the structure of the latent variable $\bm z$. 

\subsection{Variational Autoencoder (VAE)}
\label{ssec:vae}
\begin{figure}[!ht]
	\centering
	\includegraphics[trim = 5mm 40mm 5mm 30mm, clip, width = 1.0\textwidth]{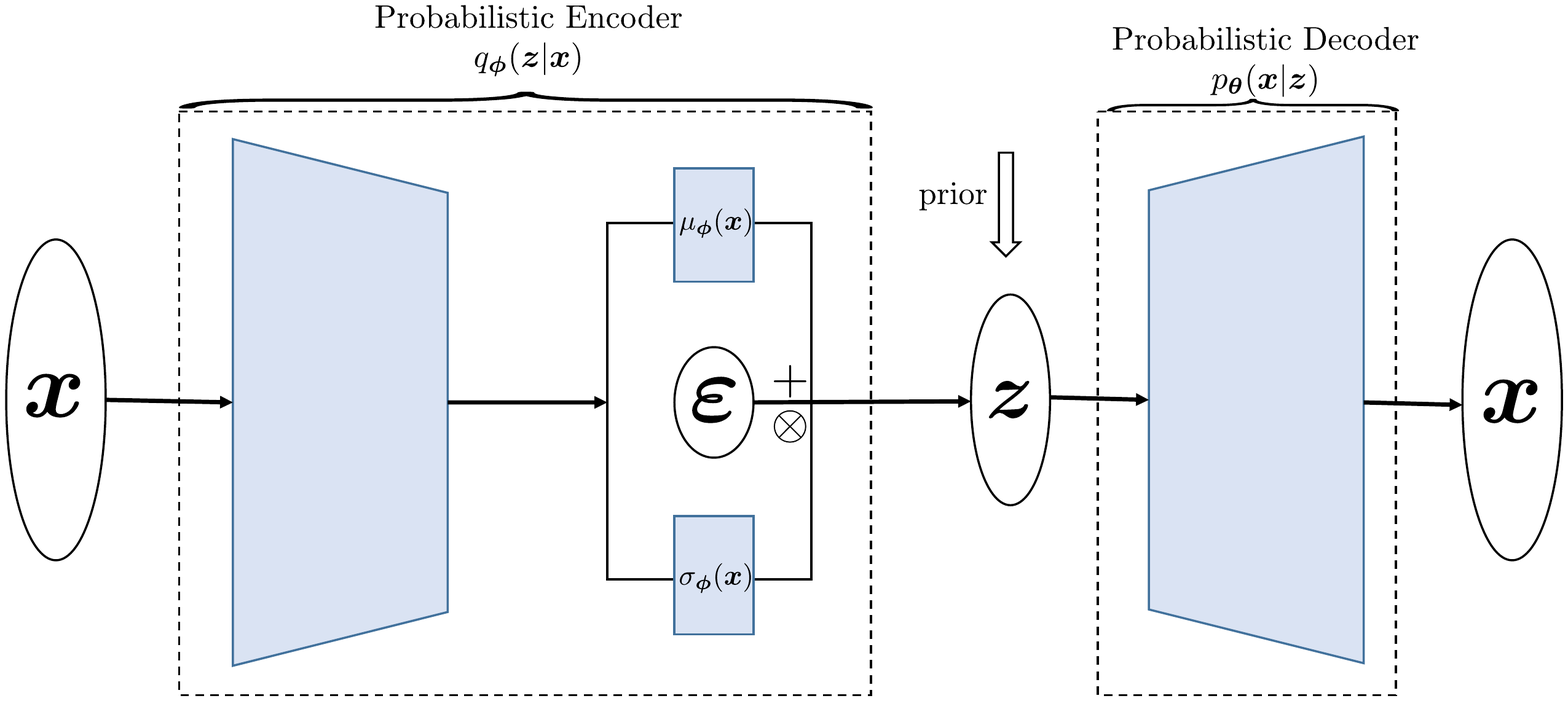}
	\caption{The probabilistic encoder $q_{\bm\phi}(\bm z\vert\bm x)$ of a VAE produces two separate vectors, $\bm\mu_{\bm\phi}(\bm x)$ and $\bm\sigma_{\bm\phi}(\bm x)$, which respectively represent the mean and standard deviation of resulting latent variable $\bm z$ following a multivariate Gaussian distribution. The random vector $\bm\varepsilon\sim\mathcal{N}(\bm 0,\bm I)$ provides randomness for the encoder output $\bm z$ and is used for the reparameterization trick in Equation~\eqref{eq:repara2}.}
	\label{fig:vae}
\end{figure}

This section introduces VAE \cite{kingma14auto,rezende2014stochastic}, a widely-used deep generative model capable of using samples of $\bm x$ to construct an estimate of $p(\bm x)$ from which new samples of $\bm x$ can be drawn. 
As a deep Bayesian model, VAE compresses and reconstructs data in a non-linear and probabilistic manner, while regularizing the model via a Kullback–Leibler (KL) divergence term (Equation~\eqref{eq:elbo}, which distinguishes it from regular AE models. 
The VAE is composed of two distinct probabilistic components, namely the encoder and the decoder as depicted in Figure~\ref{fig:vae}. 
In contrast to AEs, the encoder and decoder of a VAE map data to random vectors, rather than deterministic values. 
The probabilistic encoder produces two separate vectors, representing the mean and standard deviation of a resulting multivariate Gaussian random vector $\bm z$. 
In this context, the covariance matrix of $\bm z$ is assumed to be diagonal.
The probabilistic decoder maps the latent variable $\bm z$ back to the observation space by sampling from the decoder's output distribution. 
When the expected output is a continuous random variable, which is the primary focus of this work, the decoder result is traditionally assumed to be deterministic and returns the mean value of the decoder distribution \citep{kingma14auto}. 
In other words, the decoder $p(\bm x\vert\bm z)$ becomes a Dirac distribution located at $D(\bm z)$, where $D$ is the deterministic decoder function. 
By enforcing a prior on the latent variable $\bm z$, the VAE can synthesize new samples of $\bm x$ by sampling the latent variable and evaluating the decoder.

In detail, VAE introduces a latent variable $\bm z$ with its prior $p(\bm z)$ in a low-dimensional latent space and parameterizes a probabilistic decoder $p_{\bm\theta}(\bm x\vert\bm z)$ with parameters $\bm\theta$ to establish a joint pdf $p_{\bm\theta}(\bm x,\bm z)$. 
According to the Bayes' rule, the posterior density is given by
\begin{align}
\label{eq:vae-post}
    p_{\bm\theta}(\bm z\vert\bm x) = \frac{p(\bm z)p_{\bm\theta}(\bm x\vert\bm z)}{\int p_{\bm\theta}(\bm x,\bm z)d\bm z}.
\end{align}
In practice, computing $p_{\bm\theta}(\bm z\vert\bm x)$ is intractable due to the unknown marginal density $\int p_{\bm\theta}(\bm x,\bm z)d\bm z$. 
To address this issue, VAE employs a variational inference approach \cite{blei2017variational} and approximates the posterior density with a pdf $q_{\bm\phi}(\bm z\vert\bm x)$ parameterized by $\bm\phi$. 
By introducing the variational replacement $q_{\bm\phi}$, the log-likelihood of $\bm x$ can be decomposed as 
\begin{align}
\label{eq:lkh-elbo}
    \log(p_{\bm\theta}(\bm x)) &= \KL\big(q_{\bm\phi}(\bm z|\bm x)||p_{\bm\theta}(\bm z|\bm x))\big) + \underbrace{\mathbb{E}_{q_{\bm\phi}}\log(\frac{p_{\bm\theta}(\bm x,\bm z)}{q_{\bm\phi}(\bm z|\bm x)})}_{\ELBO},
\end{align}
where $\KL(\cdot\|\cdot)$ is the Kullback-Leibler (KL) divergence and $\E_{q_{\bm\phi}}$ is the expectation over $q_{\bm\phi}(\bm z\vert\bm x)$. 
The KL divergence term in Equation~\eqref{eq:lkh-elbo} measures the discrepancy between the true posterior $p_{\bm\theta}(\bm z\vert\bm x)$ and the variational posterior $q_{\bm\phi}(\bm z\vert \bm x)$ and is unknown in practice. 
The second term in Equation~\eqref{eq:lkh-elbo} is known as evidence lower bound (ELBO), which is a lower bound of the log-likelihood due to the non-negativity of the KL divergence.  
In variational inference, ELBO is maximized instead of the log-likelihood due to its tractable form. 
By maximizing ELBO, the VAE model enhances a lower bound value of the log-likelihood and mitigates the discrepancy between the variational and true posteriors. 

The VAE objective function, ELBO, can be further decomposed into two parts
\begin{align}
\label{eq:elbo}
    \text{ELBO}(\bm\phi,\bm\theta) &= \mathbb{E}_{q_{\bm\phi}}\log(\frac{p_{\bm\theta}(\bm x,\bm z)}{q_{\bm\phi}(\bm z|\bm x)})\\
                &=  - \underbrace{\KL(q_{\bm\phi}(\bm z|\bm x)||p(\bm z))}_{\textup{regularization term}} + \underbrace{\E_{q_{\bm\phi}}\log(p_{\bm\theta}(\bm x|\bm z))}_{\textup{reconstruction term}}.
\end{align}
The first part is the KL divergence between the prior $p(\bm z)$ and the variational posterior $q_{\bm\phi}(\bm z|\bm x)$ measuring the distance between the two densities. 
The second term, $\mathbb{E}_{q_{\bm\phi}}\log(p_{\bm\theta}(\bm x|\bm z))$, is the conditional log-likelihood of $\bm x$ that is averaged over the variational posterior $\bm z\sim q_{\bm\phi}$. 
This component is often perceived as a negative reconstruction error. 
For example, when the conditional density $p_{\bm\theta}(\bm x|\bm z)$ is Gaussian centered at the decoder output $D_{\bm\theta}(\bm z)$, where $D_{\bm\theta}(\bm z)$ is a neural-network-based decoder function, $\log(p_{\bm\theta}(\bm x|\bm z))$ becomes the negative 2-norm reconstruction error $-\lVert \bm x - D_{\bm\theta}(\bm z)\rVert^2$. 

In order to estimate $\bm\theta$ and $\bm\phi$, gradient ascent is applied to maximize ELBO with gradients $\nabla_{\bm\phi}\ELBO$ and $\nabla_{\bm\theta}\ELBO$.
However, the gradient of ELBO with respect to $\bm\phi$, i.e.,
\begin{equation}
\label{eq:repara1}
    \nabla_{\bm\phi} \text{ELBO}= \nabla_{\bm\phi} \mathbb{E}_{q_{\bm\phi}}\log(\frac{p_{\bm\theta}(\bm x,\bm z)}{q_{\bm\phi}(\bm z|\bm x)})
\end{equation}
cannot be computed directly since the expectation $\E_{q_{\bm\phi}}$ depends on $\bm\phi$. 
Instead, VAE uses a new random vector $\bm\varepsilon\sim\mathcal{N}(0,\bm I)$ and represents latent samples as $\bm z_{\bm\varepsilon} = \bm\sigma_{\bm\phi}(\bm x)\otimes\bm\varepsilon+\bm\mu_{\bm\phi}(\bm x)$, where $\otimes$ is Hadamard (element-wise) product. 
Stochastic gradient ascent (or its variants) is performed for each mini-batch of samples $\{\bm x_i\}_{i=1}^B$
by passing them through the encoder and obtaining $\sigma_{\bm\phi}(\bm x_i)$ and $\bm\mu_{\bm\phi}(\bm x_i)$, and generating new $\bm z_{\bm\varepsilon_i}$ by sampling $\bm\varepsilon_i$.
An unbiased estimate of the gradient is generated via
\begin{equation}
\label{eq:repara2}
\begin{aligned}
    \nabla_{\bm\phi} \text{ELBO} &= \nabla_{\bm\phi} \mathbb{E}_{\bm \varepsilon}\log(\frac{p_{\bm\theta}(\bm x\vert\bm z_{\bm\varepsilon})p(\bm z_{\bm\varepsilon})}{q_{\bm\phi}(\bm z_{\bm\varepsilon}|\bm x)})\\
                            &=  \mathbb{E}_{\bm \varepsilon}\nabla_{\bm\phi}\log(\frac{p_{\bm\theta}(\bm x\vert\bm z_{\bm\varepsilon})p(\bm z_{\bm\varepsilon})}{q_{\bm\phi}(\bm z_{\bm\varepsilon}|\bm x)})\\
                            &\approx \frac{1}{B}\sum_{i=1}^B\nabla_{\bm\phi}\log(\frac{p_{\bm\theta}(\bm x=\bm x_i\vert\bm z=\bm z_{\bm\varepsilon_i})p(\bm z=\bm z_{\bm\varepsilon_i})}{q_{\bm\phi}(\bm z=\bm z_{\bm\varepsilon_i}|\bm x=\bm x_i)}).
\end{aligned}
\end{equation}
The method for estimating the gradient in Equation~\eqref{eq:repara2}, known as the reparametrization trick \cite{kingma14auto}, can be applied to any form of $q_{\bm\phi}(\bm z\vert \bm x)$, provided that it is associated with an easy-to-sample distribution. 
It further allows for decoupling of the expectation from $\bm\phi$ in Equation~\eqref{eq:repara2},  thereby enabling the optimization of the objective function. 

\subsection{Auto-regressive Method}
\label{ssec:auto-regressive}
The central challenge of bi-fidelity modeling is to establish a connection between LF and HF model outputs. 
The VAE in Section~\ref{ssec:vae} provides a solution for building the encoder and decoder for searching an appropriate latent space. When using bi-fidelity data, we additionally require a suitable architecture for the latent variable $\bm z$ to model the relation between the LF and HF solutions. 
This architecture must be relatively simple as we assume only limited HF data is available. 
For example, in \cite{chen2023feature}, the authors use an encoder-decoder structure in conjunction with a latent bi-fidelity modeling approach that minimizes the distance between the reduced basis coefficients. 
We extend this method to a more general form. 

For the case of the probabilistic encoder and decoder, we split the latent random vector $\bm z$ into two parts, $\bm z^L$ and $\bm z^H$, and apply a linear auto-regression from $\bm z^L$ to $\bm z^H$, inspired by the well-known Gaussian process (GP) based linear auto-regressive method \cite{kennedy2000predicting,kennedy2001bayesian,le2013multi}. 
This approach incorporates multivariate Gaussian priors for both fidelity models and postulates a linear, element-wise relationship between the models. 
The HF latent random vector $\bm z^H$ can be represented as a transformation of the LF latent random vector $\bm z^L$ through 
\begin{equation}
\label{eq:co-kriging}
    z^H_i = a_i z^L_i+b_i, \forall i=1,2,\dots,d
\end{equation}
where $a_i$ serves as a scaling factor and $b_i$ is a Gaussian random variable. 
In some works, e.g., \cite{kennedy2001bayesian,kennedy2000predicting}, $z^H_i$ and $z^L_i$ are indexed with a spatial variable, which has been omitted here for clarity. 
The model assumes that no knowledge of $z^H_i$ can be extracted from $z^L_j$ if $z^L_i$ is known and $i\neq j$, which implies $\text{Cov}(z^H_i,z^L_j|z^L_i)=0, \forall i\neq j$. 

\section{Bi-fidelity Variational Auto-encoder (BF-VAE)}
\label{sec:methodology}
In this section, we present the BF-VAE model. 
Section~\ref{ssec:main-alg} outlines the architecture of BF-VAE, a bi-fidelity extension of the ELBO objective function, and an algorithm designed to train BF-VAE.
The bi-fidelity information bottleneck (BF-IB) theory is introduced in Section~\ref{ssec:bf-vib}, providing an interpretation of BF-VAE from the perspective of information theory.
Section~\ref{ssec:error-analysis} delves into the analysis of an error stemming from the probabilistic encoder trained by LF data.

\subsection{Architecture, Objective Functions, and Algorithm} 
\label{ssec:main-alg} 
\begin{figure}[!ht]
	\centering
	\includegraphics[trim = 5mm 25mm 30mm 10mm, clip, width = 1.0\textwidth]{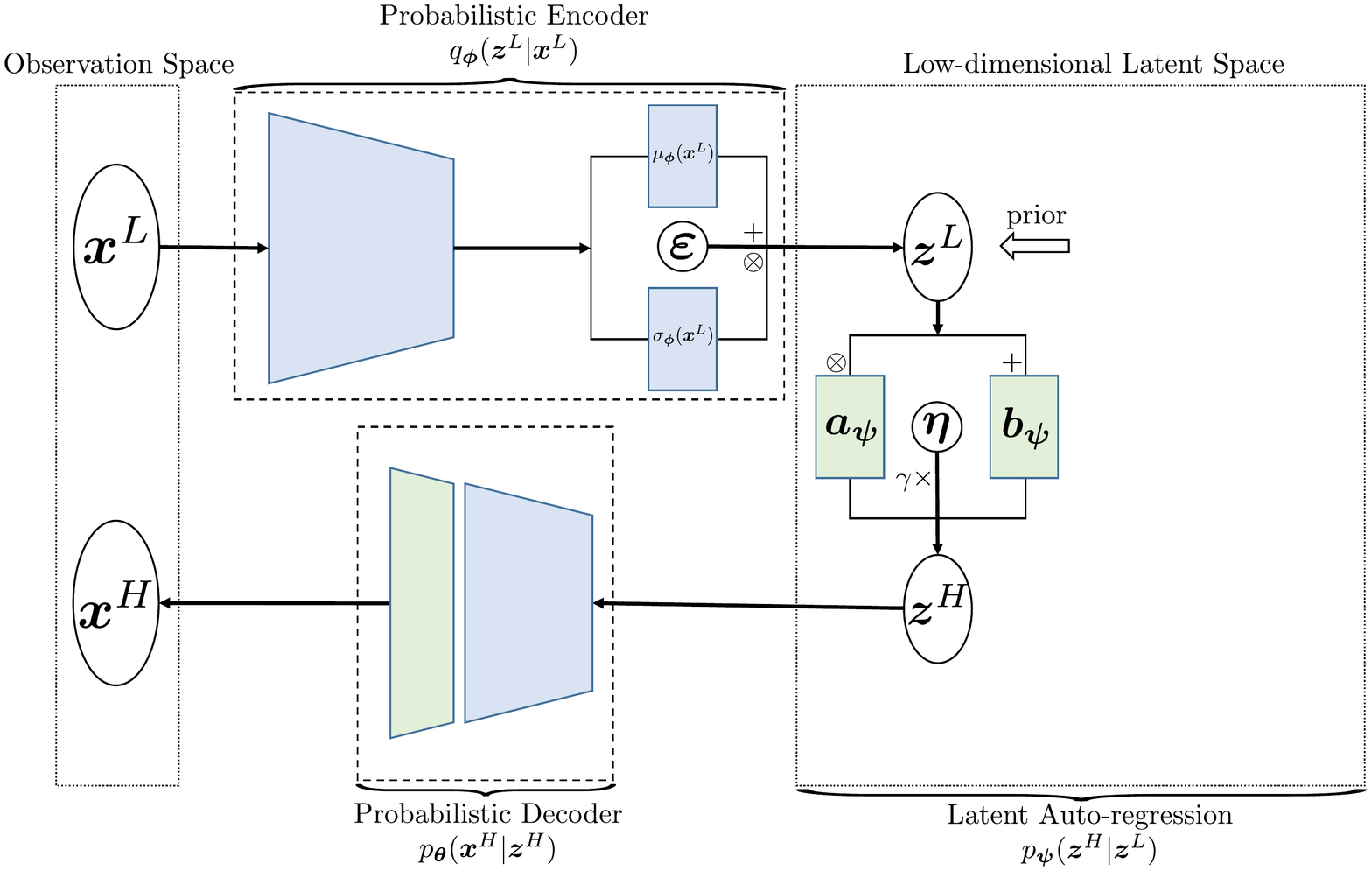}
	\caption{Structure of the proposed BF-VAE model. The probabilistic encoder $q_{\bm\phi}(\bm z^L\vert\bm x^L)$ produces two independent vectors, $\mu_{\bm\phi}(\bm x^L)$ and $\sigma_{\bm\phi}(\bm x^L)$, which represent the mean and standard deviation of a resulting multivariate Gaussian. The latent auto-regression $p_{\bm\psi}(\bm z^H\vert\bm z^L)$ is a simplified single-layer neural network $\bm K_{\bm\psi}$ defined in Equation~\eqref{eq:latent-layer-K} added with a noise $\gamma\bm\eta$. The probabilistic decoder $p_{\bm\theta}(\bm x^H\vert\bm z^H)$ is pre-trained by LF data via the transfer learning technique, with its last layer tuned by LF and HF data pairs. White circles are random vectors and colored blocks are parameterized components for training. Blue blocks are solely trained by LF data and green blocks are trained by both LF and HF data.}
	\label{fig:bfvae}
\end{figure}

The principle behind BF-VAE involves maximizing a lower bound of the HF log-likelihood, as VAE does, but primarily utilizing LF data.
To achieve this, a VAE-based structure is devised to leverage a latent space to model the relationship between the LF and HF data.
The BF-VAE model is comprised of three probabilistic components: an encoder, a latent auto-regression, and a decoder.
The probabilistic encoder $q_{\bm\phi}(\bm z^L\vert\bm x^L)$, parameterized with $\bm\phi$ and trained using LF data, maps LF observations into LF latent representations.
The latent auto-regression $p_{\bm\psi}(\bm z^H\vert\bm z^L)$, parameterized with $\bm\psi$ and specifically designed for a bi-fidelity regression in the latent space, as shown in Equation~\eqref{eq:latent-layer-K}, significantly reduces the amount of HF data required for training due to its low-dimensionality.
The probabilistic decoder $p_{\bm\theta}(\bm x^H\vert\bm z^H)$, parameterized with $\bm\theta$, is first pre-trained with LF data and then refined with HF data, mapping the HF latent representations back into the observation space by returning the mean of the resulting HF distribution.
A schematic illustration of the proposed BF-VAE model is depicted in Figure~\ref{fig:bfvae}.

A crucial part of BF-VAE is building a connection from the LF latent variable $\bm z^L$ to the HF latent variable $\bm z^H$.
As presented in Section~\ref{ssec:auto-regressive}, we specify the latent conditional density $p_{\bm\psi}(\bm z^H\vert\bm z^L)$ to be a linear auto-regressive model, assumed to follow the Gaussian distribution $\mathcal{N}(\bm K_{\bm\psi}(\bm z^L),\gamma^2\bm I)$ with parameters $\bm\psi$. 
The mapping $\bm K_{\bm \psi}$ consists of two parameterized vector components, $\bm a_{\bm\psi}$ and $\bm b_{\bm\psi}$, defined by the affine transformation
\begin{equation}
\label{eq:latent-layer-K}
    \bm K_{\bm\psi}(\bm z^L) = \bm a_{\bm\psi}\otimes\bm z^L + \bm b_{\bm\psi}.
\end{equation}
In this work, $\bm K_{\bm\psi}$ is implemented as a simplified single-layer neural network with a diagonal weight matrix and a bias vector. 
The hyperparameter $\gamma$ is fixed for all entries for simplicity. 
When $\gamma\to 0$, $p_{\bm\psi}(\bm z^H\vert\bm z^L)$ converges in distribution to the Dirac distribution $\delta_{\bm K_{\bm\psi}(\bm z^L)}$, which makes the latent auto-regression a deterministic map. 
The hyperparameter $\gamma$ represents our confidence on how accurately $\bm K_{\bm\psi}$ captures the relation between the LF and HF latent variables; see more discussion about $\gamma$ in Section~\ref{ssec:hyper-tune}.

The objective function of BF-VAE is a variational lower bound of the HF log-likelihood as follows
\begin{equation}
\label{eq:bf-lkh}
\begin{aligned}
    \log p_{\bm\theta,\bm\psi}(\bm x^H) &= \KL\big(q_{\bm\phi}(\bm z_{\bm\psi}\vert\bm x^L)\|p_{\bm\theta}(\bm z_{\bm\psi}\vert\bm x^H)\big) + \mathbb{E}_{q_{\bm\phi}(\bm z_{\bm\psi}\vert \bm x^L)}\bigg[\log(\frac{p_{\bm \theta}(\bm x^H,\bm z_{\bm\psi})}{q_{\bm\phi}(\bm z_{\bm\psi}\vert\bm x^L)})\bigg]\\
    &\geq \mathbb{E}_{q_{\bm\phi}(\bm z_{\bm\psi}\vert \bm x^L)}\bigg[\log(\frac{p_{\bm \theta}(\bm x^H,\bm z_{\bm\psi})}{q_{\bm\phi}(\bm z_{\bm\psi}\vert\bm x^L)})\bigg] = \ELBO^{\BF}(\bm\phi,\bm\psi,\bm\theta),
\end{aligned}
\end{equation}
where the pdf of $\bm z_{\bm\psi}\coloneqq(\bm z^L,\bm z^H)$ is determined by the latent conditional density $p_{\bm\psi}(\bm z^H\vert\bm z^L)$ and the prior $p(\bm z^L)$. 
The above inequality follows from the non-negativity property of KL divergence. 
The lower bound of thw HF log-likelihood in Equation~\eqref{eq:bf-lkh} is called the bi-fidelity ELBO (BF-ELBO), and denoted as $\ELBO^{\BF}(\bm\phi,\bm\psi,\bm\theta)$. 
The BF-ELBO consists of two terms, namely
\begin{align}
\label{eq:bf-elbo}
    \ELBO^{\BF}(\bm\phi,\bm\psi,\bm\theta)    &= -\underbrace{\KL\big(q_{\bm\phi}(\bm z^L\vert\bm x^L)\|p(\bm z^L)\big)}_{\textup{regularization term}} + \underbrace{\mathbb{E}_{q_{\bm\phi}(\bm z_{\bm\psi}\vert \bm x^L)}\big[\log p_{\bm\theta}(\bm x^H\vert \bm z_{\bm\psi})\big]}_{\textup{HF reconstruction term}}.
\end{align}
The first term regularizes the encoder training by enforcing the encoder output to be close to the prior $p(\bm z^L)$. 
The second term is the HF log-likelihood conditioned on the latent variable $\bm z_{\bm\psi}$ and perceived as the HF reconstruction term. 
For example, when $p_{\bm\theta}(\bm x^H\vert\bm z_{\bm\psi})$ is a Gaussian centered at the decoder output $D_{\bm\theta}(\bm z^H)$ with covariance $\beta\bm I$, the HF reconstruction term is a negative 2-norm $-\beta^{-1}\lVert \bm x^H - D_{\bm\theta}(\bm z^H)\rVert^2$ with $\bm z^H$ drawn from the encoder and the latent auto-regression with input $\bm x^L$. 
Note that by the Markov property, $p(\bm x^H\vert\bm z_{\bm\psi})$ is equivalent to $p(\bm x^H\vert\bm z^H)$.
We use $\bm z_{\bm\psi}$ as the conditional variable for $p_{\bm\theta}(\bm x^H\vert \bm z_{\bm\psi})$ so that it is consistent with the expectation $\E_{q_{\bm\phi}(\bm z_{\bm\psi}\vert \bm x^L)}$.
A detailed derivation of Equations~\eqref{eq:bf-lkh} and \eqref{eq:bf-elbo} are presented in \ref{apdx:bf-elbo}.

Optimizing BF-ELBO requires a large amount of both LF and HF data from their joint distribution $p(\bm x^L,\bm x^H)$ for convergence.
However, the scarcity of HF data presents a challenge under the bi-fidelity setting. 
To address this issue, we apply a transfer learning technique, in which we opt to train the encoder and decoder using a large set of LF data, considering that the parameter spaces of $\bm\phi$ and $\bm\theta$ are significantly larger than that of $\bm\psi$. 
The small parameter space of $\bm\psi$ as a single layer in the low-dimensional latent space allows it to be trained solely with pairs of LF and HF data.
As a result, we optimize the BF-ELBO in two steps, with two separate objectives, 
\begin{align}
\label{eq:lf-elbo}
    \ELBO^{\LF}(\bm\phi,\bm\theta) &= -\KL\big(q_{\bm\phi}(\bm z^L\vert\bm x^L)\|p(\bm z^L)\big) + \mathbb{E}_{q_{\bm\phi}(\bm z^L\vert\bm x^L)}\big[\log(p_{\bm\theta}(\bm x^L|\bm z^L))\big],\\
    \ELBO^{\HF}(\bm\psi,\bm\theta) &= \mathbb{E}_{q_{\bm\phi^{L*}}(\bm z_{\bm\psi}\vert \bm x^L)}\big[\log p_{\bm\theta}(\bm x^H\vert \bm z_{\bm\psi})\big],
    \label{eq:hf-elbo}
\end{align}
where $\bm\phi^{L*}$ in $\ELBO^{\HF}$ is the optimal $\bm\phi$ for maximizing $\ELBO^{\LF}$. 
The first objective function $\ELBO^\LF(\bm\phi,\bm\theta)$ is equivalent to a regular ELBO function discussed in Equation~\eqref{eq:elbo}, as it trains a low-fidelity VAE (LF-VAE) solely using LF data. 
The trained LF-VAE returns the optimal LF encoder parameters $\bm\phi^{L*}$ and LF decoder parameters $\bm\theta^{L*}$. 
We assume the optimal HF decoder parameters $\bm\theta^{H*}$ is close to $\bm\theta^{L*}$ in the parameter space. 
Furthermore, we fix the decoder's parameters except for the last layer and set $\bm\theta^{L*}$ as the initial value for further optimizing $\ELBO^\HF$ using both LF and HF data to obtain optimal HF parameters $\bm\psi^{H*}, \bm\theta^{H*}$. 
Note that $\bm\theta^{H*}$ and $\bm\theta^{L*}$ are the same except for entries corresponding to the decoder's last layer, due to this transfer learning technique. 

The presence of the parameter $\bm\psi$ in the expectation term in Equation~\eqref{eq:hf-elbo} poses a challenge for the estimation of the gradients with respect to $\bm\psi$. 
To address this, we leverage the reparameterization trick outlined in Equation~\eqref{eq:repara2}. 
Specifically, we introduce an auxiliary vector $\bm\eta\sim\mathcal{N}(\bm 0,\bm I)$ and set
\begin{align}
\label{eq:repara3}
    \bm z^H_{\bm\eta} = \gamma\bm\eta +\bm K_{\bm\psi}(\bm z^L).
\end{align}
With mini-batch bi-fidelity samples $\{\bm x^L_i,\bm x^H_i\}_{i=1}^B$, the gradient w.r.t.\ $\bm\psi$ is estimated as
\begin{equation}
\label{eq:hf-elbo-repara}
\begin{aligned}
    \nabla_{\bm\psi} \ELBO^{\HF}(\bm\psi,\bm\theta)  &= \mathbb{E}_{p_{\bm\phi^{L*}}(\bm z^L\vert\bm x^L)} [\nabla_{\bm\psi}\mathbb{E}_{p_{\bm\psi}(\bm z^H\vert\bm z^L)} [\log p_{\bm\theta}(\bm x^H\vert\bm z^H)]]\\
                            &= \mathbb{E}_{p_{\bm\phi^{L*}}(\bm z^L\vert\bm x^L)} [\mathbb{E}_{\bm\eta}[\nabla_{\bm\psi} \log p_{\bm\theta}(\bm x^H\vert\bm z^H_{\bm\eta})]]\\
                            &\approx \frac{1}{B}\sum_{i=1}^B\nabla_{\bm\psi} \log p_{\bm\theta}(\bm x^H=\bm x^H_i\vert\bm z^H=\bm z^H_{\bm\eta_i}),
\end{aligned}
\end{equation}
where $B$ is batch size and $\bm z^H_{\bm\eta_i}$ is the $i$-th sample from $\bm x^L_i$ and $\bm\eta_i$ as shown in Equation~\eqref{eq:repara3}.
Using the estimated gradients, we maximize $\ELBO^{\LF}(\bm\phi,\bm\theta)$ and $\ELBO^{\HF}(\bm\psi,\bm\theta)$ via stochastic gradient ascent (or its variants). 
To synthesize HF QoI samples, we sample from $p(\bm z^L)$ and subsequently propagate the samples through the trained latent auto-regressor with parameters $\bm\psi^{H*}$ and subsequently the decoder with parameters $\bm\theta^{H*}$.
A summary of the steps in BF-VAE is provided in Algorithm~\ref{alg:BFVAE}.

\begin{algorithm}[ht]
		\DontPrintSemicolon
		\caption{Bi-Fidelity Variational Auto-Encoder (BF-VAE)\label{alg:BFVAE}}
		\KwIn{LF training set $\{\tilde{\bm x}^L_{i}\}_{i=1}^N$, LF-HF joint training set $\{(\bm x^L_{i},\bm x^H_{i})\}_{i=1}^n$}
		\KwOut{Parameters $\bm\psi^{H*},\bm\theta^{H*}$ for a HF pdf $p_{\bm\theta,\bm\psi}(\bm x^H)$}
		\begin{algorithmic}[1]
			\STATE Train a LF-VAE by maximizing $\ELBO^\LF(\bm\phi,\bm\theta)$ in Equation~\eqref{eq:lf-elbo} with LF realizations $\{\tilde{\bm x}^L_{i}\}_{i=1}^N$ to attain maximizers $\bm\phi^{L*},\bm\theta^{L*} = \argmax_{\bm\phi,\bm\theta}\ELBO^{\LF}(\bm\phi,\bm\theta).$ \label{algline:lf-trn}
			\STATE Build a BF-VAE as shown in Figure~\ref{fig:bfvae} with parameters of the encoder and the decoder assigned to be $\bm\phi^{L*},\bm\theta^{L*}$, and the latent auto-regression map $\bm K_{\bm\psi}(\cdot)$ in Equation~\eqref{eq:latent-layer-K} being initialized as an identity map. 
            \STATE Fix all the parameters of BF-VAE except the decoder's last layer and the latent auto-regression's parameters.
			\STATE Train BF-VAE by maximizing $\ELBO^{\HF}(\bm \psi, \bm\theta)$ in Equation~\eqref{eq:hf-elbo} with sample pairs $\{(\bm x^L_{i},\bm x^H_{i})\}_{i=1}^n$ and find maximizers $\bm\psi^{H*},\bm\theta^{H*} = \argmax_{\bm\psi,\bm\theta} \ELBO^{\HF}(\bm\psi,\bm\theta)$.
            \label{algline:bf-trn}
		\end{algorithmic}
\end{algorithm}
\subsection{Bi-fidelity Information Bottleneck} 
\label{ssec:bf-vib}

One of the core ideas of bi-fidelity modeling is to fully exploit information from LF data for building HF results with limited HF data. 
However, to the best of the authors' knowledge, there is no previous work that explicitly models the bi-fidelity information transfer process incorporating information theory. 
In this work, we apply the information bottleneck (IB) principle \cite{tishby2000information, shwartz2017opening} to the BF-VAE model. 
The IB principle aims to define the essence of a ``good'' latent representation of data by finding a balance between information preservation and compression. According to IB, an optimal latent representation of data is maximally informative about the output while simultaneously compressive with respect to a given input. 

In this section, we propose an interpretation of the BF-VAE model through the lens of the bi-fidelity IB (BF-IB) theory.
We show that maximizing $\ELBO^{\BF}$ in Equation~\eqref{eq:bf-elbo} is equivalent to maximizing the BF-IB objective function in Equation~\eqref{eq:bf-vib} with $\beta=1$ using $(\bm x^L,\bm x^H)$ data. 
Our analysis in this section builds a bridge between information theory and log-likelihood maximization in the bi-fidelity setting and presents a novel information-theoretic perspective on the BF-VAE model.

The mutual information, which is a non-negative, symmetric function, reflects the information that can be obtained about one random vector by observing another random vector. 
The definition of mutual information is as follows. 
\begin{definition}
    The mutual information \cite{cover1999elements} between random vectors $\bm x$ and $\bm y$ is
    \begin{equation}
    \label{eq:mutual-info-def}
        \I(\bm x,\bm y)\coloneqq \KL\big(p(\bm x,\bm y)\|p(\bm x)p(\bm y)\big) = \E_{p(\bm x,\bm y)}\log\left(\frac{p(\bm x,\bm y)}{p(\bm x)p(\bm y)}\right),
    \end{equation}
    where $p(\bm x,\bm y)$ is the joint distribution of $\bm x$ and $\bm y$. 
\end{definition}

\begin{figure}[!ht]
	\centering
	\includegraphics[trim = 0mm 45mm 0mm 55mm, clip, width = 1.0\textwidth]{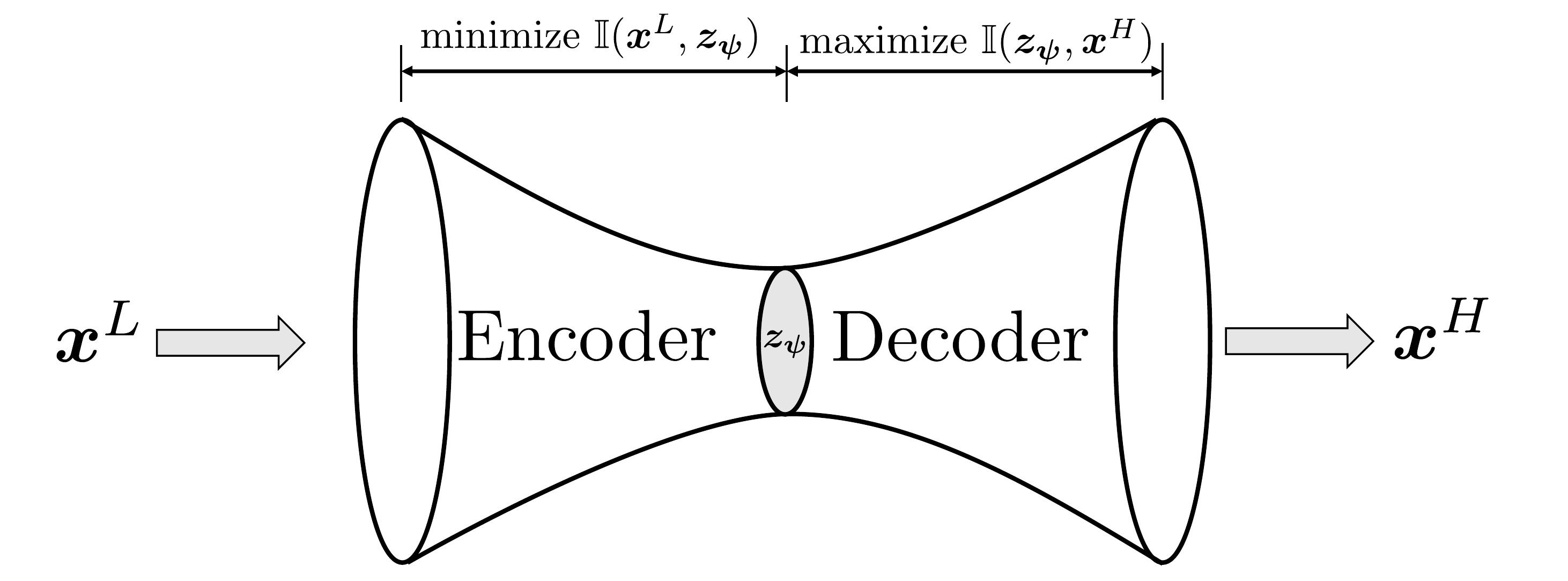}
	\caption{The bi-fidelity information bottleneck architecture has an encoder and a decoder, impacted by the information compression function $\I(\bm x^L,\bm z_{\bm\psi})$ and information preservation function $\I(\bm z_{\bm\psi},\bm x^H)$, respectively. The random vector $\bm z_{\bm\psi}$ is designed to disclose the relation between LF and HF data in the latent space. The bottleneck part is necessary since only a limited number of HF realizations are available for learning the relationship between LF and HF data.}
	\label{fig:bf-ib}
\end{figure}

In BF-VAE, our goal is to find a latent representative random vector $\bm z_{\bm\psi}$ corresponding to $\bm x^L$ for re-building HF QoI $\bm x^H$.
According to the formula (15) in \cite{tishby2000information} or formula (5.164) in \cite{murphy2023pml}, the bi-fidelity information bottleneck (BF-IB) objective function that we will maximize is
\begin{align}
\label{eq:bf-vib}
    \IB^{\BF}_\beta(\bm\phi,\bm\psi,\bm\theta) &\coloneqq \I(\bm z_{\bm\psi},\bm x^H) - \beta \I(\bm x^L,\bm z_{\bm\psi}),
\end{align}
where $\beta$ is a non-negative hyperparameter, and $\bm\phi, \bm\theta$ are parameters of the encoder and decoder, respectively. 
The first term $\I(\bm z_{\bm\psi},\bm x^H)$ represents the preserved information from $\bm z_{\bm\psi}$ to $\bm x^H$ by the decoder, while the second term $\I(\bm x^L,\bm z_{\psi})$ represents the information compressed by the encoder. 
The hyperparameter $\beta$ is adjusted to balance the tradeoff between the information compression and the preservation. 
By maximizing the BF-IB objective function, we aim to find an optimal latent random vector $\bm z_{\bm\psi}$ as well as its relation with $\bm x^L, \bm x^H$, which are parameterized by $\bm \psi, \bm \phi,$ and $\bm\theta$. 
Note that when the mutual information between $\bm x^L$ and $\bm x^H$ is zero, which means LF and HF data are independent, the searching for latent variable $\bm z_{\bm \psi}$ is vacuous.
The schematic in Figure~\ref{fig:bf-ib} describes the concept of BF-IB. 

The BF-IB objective function can be decomposed as follows,
\begin{equation}\label{eq:bf-vib-loss}
\begin{aligned}
    \IB^{\BF}_\beta(\bm\phi,\bm\psi,\bm\theta)   
    &\equiv \mathbb{E}_{p(\bm x^L,\bm x^H)}\bigg[\mathbb{E}_{q_{\bm\phi}(\bm z_{\bm\psi}\vert \bm x^L)}\big[\log p_{\bm\theta}(\bm x^H\vert \bm z_{\bm\psi})\big] - \beta\KL\big(q_{\bm\phi}(\bm z^L\vert\bm x^L)\|p(\bm z^L)\big)\bigg].
\end{aligned}
\end{equation}
When $\beta=1$, the BF-IB objective function becomes 
\begin{equation}
\begin{aligned}
    \IB^{\BF}_{\beta=1}(\bm\phi,\bm\psi,\bm\theta) 
    &\equiv \mathbb{E}_{p(\bm x^L,\bm x^H)}\bigg[\mathbb{E}_{q_{\bm\phi}(\bm z_{\bm\psi}\vert \bm x^L)}\big[\log p_{\bm\theta}(\bm x^H\vert \bm z_{\bm\psi})\big] - \KL\big(q_{\bm\phi}(\bm z^L\vert\bm x^L)\|p(\bm z^L)\big)\bigg]\\
    &=\mathbb{E}_{p(\bm x^L,\bm x^H)}[\ELBO^{\BF}(\bm\phi,\bm\psi,\bm\theta)].\\ 
\end{aligned}
\end{equation}
 This proves that the BF-IB function with $\beta=1$ is equivalent to BF-ELBO in Equation~\eqref{eq:bf-elbo} averaged with respect to the true joint distribution $p(\bm x^L,\bm x^H)$. 
The proof of \eqref{eq:bf-vib-loss} is presented in \ref{apdx:bf-ib} with the Markov property assumed. 
In Section~\ref{ssec:hyper-tune}, we incorporate the hyperparameter $\beta$ into a prior of the decoder pdf $p_{\bm\theta}(\bm x^H\vert \bm z^H)$, yielding an equivalent objective function containing $\beta$. 
Because the BF-VAE Algorithm~\ref{alg:BFVAE} approximately maximizes BF-ELBO using joint realizations from $p(\bm x^L,\bm x^H)$, it produces an output that not only maximizes a variational lower bound of the HF log-likelihood but also the IB-BF objective function.  

\subsection{Bi-fidelity Approximation Error}
\label{ssec:error-analysis}

Similar to VAE in Section~\ref{ssec:vae}, the BF-VAE model introduces an encoder to approximate the posterior $p_{\bm\theta}(\bm z_{\bm\psi}\vert\bm x^H)$, which produces an approximation error stemming from its variational form. 
Moreover, since we employ LF data as the input of the encoder, the error also depends on the similarity between LF and HF data. 
In this section, we give the form of this error and provide insight into a measurement of similarity between LF and HF data under the current Bayesian framework. 

Specifically, this error, denoted by $\Err$, is the gap between HF log-likelihood and BF-ELBO averaged with respect to the true data distribution $p(\bm x^L,\bm x^H)$. 
The BF-VAE model assigns a multivariate Gaussian distribution $q_{\bm\phi}$ to the encoder without any guarantee that the given family includes the true HF posterior. 
The error $\Err$,
\begin{align}
    \Err(\bm\psi, \bm\theta) &\coloneqq \min_{\bm\phi}\E_{p(\bm x^L,\bm x^H)}\big[\log p_{\bm\theta,\bm\psi}(\bm x^H) - \ELBO^{BF}(\bm\phi,\bm\psi,\bm\theta)\big]\\
    &= \min_{\bm\phi}\E_{p(\bm x^L,\bm x^H)}\big[\KL\big(q_{\bm\phi}(\bm z_{\bm\psi}\vert\bm x^L)\|p_{\bm\theta}(\bm z_{\bm\psi}\vert\bm x^H)\big)\big],
\end{align}
is directly derived from Equation~\eqref{eq:bf-lkh}. 
Since the error is a function of $\bm\psi$ and $\bm\theta$, the final performance of the trained BF-VAE model is determined by $\Err(\bm\psi^{H*}, \bm\theta^{H*})$, where $\bm\psi^{H*}$ and $\bm\theta^{H*}$ are the trained parameters. 
As a KL divergence averaged on the bi-fidelity data $p(\bm x^L,\bm x^H)$, the error $\Err$ can be interpreted as the average difference between the latent representations from LF and HF, which depends on the similarity between the LF and HF data. 

To improve the BF-ELBO's proximity to the HF log-likelihood, it is helpful to identify a form of $q_{\bm\phi}(\bm z_{\bm\psi}\vert\bm x^L)$ that is potentially close to $p_{\bm\theta}(\bm z_{\bm\psi}\vert\bm x^H)$. 
However, in practice, determining such a form is often infeasible \cite{shekhovtsov2021vae}. 
Alternatively, bringing LF data closer to HF data can also reduce the error by making their latent representations more similar. 

\section{Priors and Hyperparameters}
\label{sec:implementation}
In the previous section, we introduced the principle concept of the BF-VAE model.
In this section, we show two practical components of the BF-VAE model.
We discuss the prior distribution selection in Section~\ref{ssec:prior-dist}. 
An introduction to the hyperparameters and their effects on the BF-VAE performance is given in Section~\ref{ssec:hyper-tune}.  

\subsection{Choices of Prior Distributions}
\label{ssec:prior-dist}
Prior distribution, a crucial aspect of Bayesian modeling, is chosen to reflect our prior belief of the parameter or facilitate computation.  All the prior distributions utilized in the BF-VAE model are outlined in Table~\ref{table:priors}.

\begin{table}[ht]
	\centering
	\caption{
		Prior selection for different components of LF-VAE and BF-VAE is discussed in Section~\ref{ssec:main-alg}. Here, $\bm\mu_{\bm\phi}(\bm x^L)$ and $\bm\sigma_{\bm\phi}(\bm x^L)$ are the outputs of the variational encoder. $\bm K_{\bm\psi}$ is a parameterized latent mapping in Equation~\eqref{eq:latent-layer-K}. $\{\gamma,\beta\}$ are hyperparameters.
	}
    \begin{tabular*}{\linewidth}{@{\extracolsep{\fill}} lccc }
	\toprule
	  Component & Notation  & VAE Model(s) & Prior Distribution  \\
	\midrule
	LF Latent Variable &$p(\bm z^L)$ & LF-VAE, BF-VAE & $\mathcal{N}(\bm 0,\bm I)$ \\
	Variational Encoder &$q_{\bm \phi}(\bm z^L\vert\bm x^L)$ 	&  LF-VAE, BF-VAE	& $\mathcal{N}(\bm\mu_{\bm\phi}(\bm x^L),\bm\sigma_{\bm\phi}(\bm x^L))$\\
    Latent Auto-regression &$p_{\bm\psi}(\bm z^H\vert\bm z^L)$ 	&  BF-VAE	& $\mathcal{N}(\bm K_{\bm\psi}(\bm z^L),\gamma^2 \bm I)$\\
    LF Decoder & $p_{\bm\theta}(\bm x^L\vert\bm z^L)$ 	&  LF-VAE	& $\mathcal{N}(D_{\bm\theta}(\bm z^L),\beta\bm I)$\\
    HF Decoder & $p_{\bm\theta}(\bm x^H\vert\bm z^H)$ 	&  BF-VAE	& $\mathcal{N}(D_{\bm\theta}(\bm z^H),\beta\bm I)$\\
	\bottomrule
    \end{tabular*}
	\label{table:priors}
\end{table}

Let $\{\tilde{\bm x}^L_{i}\}_{i=1}^N\sim p(\bm x^L)$ and $\{\bm x^L_{j}, \bm x^H_{j}\}_{j=1}^n\sim p(\bm x^L, \bm x^H)$ denote the LF and BF training datasets, respectively.
When using the priors in Table~\ref{table:priors}, the LF-ELBO in Equation~\eqref{eq:lf-elbo} becomes
\begin{align}
\label{eq:lf-elbo-new}
    \ELBO^{\LF}_\beta(\bm\phi,\bm\theta) &= 
    \underbrace{-\frac{1}{2}\big(\lVert\bm\mu_{\bm\phi}( \tilde{\bm x}^L_{i})\rVert^2_2+\lVert\bm\sigma_{\bm\phi}(\tilde{\bm x}^L_{i})\rVert^2_2-\bm 1^T\log \bm\sigma^2_{\bm\phi}(\tilde{\bm x}^L_{i})\big)}_\textup{regularization}\\
    &\quad +\frac{1}{N}\sum_{i=1}^N\bigg[\underbrace{\beta^{-1}\mathbb{E}_{q_{\bm\phi}(\bm z^L\vert \bm x^L = \tilde{\bm x}^L_{i})}\lVert D_{\bm\theta}(\bm z^L) - \bm x^L_{i}\rVert_2^2}_\textup{LF reconstruction}\bigg].
\end{align}
Similarly, the HF-ELBO in Equation~\eqref{eq:hf-elbo} becomes 
\begin{align}
    \ELBO^{\HF}_\beta(\bm\psi,\bm\theta) &= \frac{1}{n}\sum_{i=1}^n\bigg[\underbrace{\beta^{-1}\mathbb{E}_{p_{\bm\phi^{L*}}(\bm z^L\vert \bm x^L = \bm x^L_{i})} \big[\lVert D_{\bm\theta}(\bm z^H_{\bm\eta_i}) - \bm x^H_{i}\rVert_2^2\big]}_\textup{BF reconstruction}\bigg],
    \label{eq:hf-elbo-new}
\end{align}
where $\bm z^H_{\bm\eta_i}$ is computed as in Equation~\eqref{eq:repara3} with $\bm\eta_i\sim\mathcal{N}(\bm 0,\bm I)$. 

\subsection{Hyperparameter Setting}
\label{ssec:hyper-tune}
BF-VAE consists of two primary hyperparameters, namely $\beta$ in Equation~\eqref{eq:bf-vib-loss} and Table~\ref{table:priors} and $\gamma$ in Equation~\eqref{eq:repara3}, which must be specified prior to training. Note that $\ELBO^\BF(\bm\phi,\bm\psi,\bm\theta)$ in Equation~\eqref{eq:bf-elbo} with priors outlined in Table~\ref{table:priors} is 
\begin{align}
    \ELBO^{\BF}(\bm\phi,\bm\psi,\bm\theta)    &= -\KL\big(q_{\bm\phi}(\bm z^L\vert\bm x^L)\|p(\bm z^L)\big)+\beta^{-1}\mathbb{E}_{q_{\bm\phi}(\bm z_{\bm\psi}\vert \bm x^L)}\big[\lVert \bm x^H - D_{\bm\theta}(\bm z^H)\rVert^2\big] \\
    &\equiv -\beta\KL\big(q_{\bm\phi}(\bm z^L\vert\bm x^L)\|p(\bm z^L)\big)+\mathbb{E}_{q_{\bm\phi}(\bm z_{\bm\psi}\vert \bm x^L)}\big[\lVert \bm x^H - D_{\bm\theta}(\bm z^H)\rVert^2\big] ,
\end{align}
where $\beta$ is a hyperparameter adjusting the contribution of the KL regularization term and also aligns with the $\beta$ in Equation~\eqref{eq:bf-vib-loss}. 
Thus, the parameter $\beta$ in the decoder prior of Table~\ref{table:priors} is analogous to the one in the BF-IB objective function in Equation~\eqref{eq:bf-vib}, which also plays a similar role to the $\beta$ parameter in $\beta$-VAE \cite{higgins2017betavae}. 
As discussed in Section~\ref{ssec:bf-vib}, the value of $\beta$ balances the tradeoff between the information compression and preservation from the perspective of IB and may be derived from prior knowledge or may be tuned using the validation error of the LF-VAE model. 

Following the discussion of Section~\ref{ssec:main-alg}, the hyperparameter $\gamma$ serves as the variance of the latent auto-regressive model. 
It indicates the degree of confidence in the accuracy of $\bm K_{\bm\psi}$ (defined in Equation~\eqref{eq:latent-layer-K}) when  modeling the latent variables of the LF and HF models. 
A larger value of $\gamma$ allows the auto-regression output to deviate further from $\bm K_{\bm\psi}$ but also increases the variance of the ELBO gradients due to a more noisy reparameterized $\bm z_{\bm\varepsilon}^H$, as shown in Equation~\eqref{eq:repara3}. 
In our numerical experiments, we observe that linear auto-regression in the latent space is capable of accurately capturing the relationship between the LF and HF latent representations, which means that we are able to choose $\gamma$ to be small.
Since a smaller $\gamma$ ensures faster convergence when optimizing the HF-ELBO, we 
therefore choose $\gamma = 0$. 

\section{Empirical Results}
\label{sec:experiments}
In this section, we present empirical results obtained by applying BF-VAE to three PDE-based forward UQ problems. In more details, we first simulate a composite beam in Section~\ref{ssec:beam}, then we discuss studying a thermally-driven cavity fluid flow with a high-dimensional uncertain input in Section~\ref{ssec:cavity}. Finally, we consider a 1D viscous Burgers' equation in Section~\ref{ssec:burgers}.
For each problem, we present the computational cost ratio between the HF and LF models. The outcomes of the BF-VAE model are then compared with HF-VAE, a standard VAE model  trained exclusively with high-fidelity data. These two models have the same architecture and activation functions.

Our primary objective is to showcase the efficacy of BF-VAE in improving the accuracy of VAE models trained using high-fidelity training data only, particularly when limited high-fidelity data is available. 
\footnote{The code implementation is available at \href{https://github.com/CU-UQ/Bi-fidelity-VAE}{https://github.com/CU-UQ/Bi-fidelity-VAE}.}

To examine the quality of data produced by a generative model, it is crucial to use an appropriate evaluation metric.
While human evaluation may be adequate for determining the quality of outputs from models generating images and text, such an approach is not generally appropriate for evaluating the quality generated data corresponding to PDE solutions.
Therefore, we seek to identify a statistical distance that allows us to compare the difference between the true $p(\bm x^H)$ and the VAE surrogates $p_{\bm\psi,\bm\theta}(\bm x^H)$ without incurring excessive computational cost. 
For deep generative models, there are two major evaluation options: Frechet inception distance (FID) \cite{heusel2017gans} and kernel inception distance (KID) \cite{binkowski2018demystifying}.
FID is most appropriate for evaluating image-based generative models as it uses a pre-trained convolutional neural network \cite{szegedy2016rethinking}. In this study, 
we employ KID, which stems from a statistical distance named maximum mean discrepancy (MMD) \cite{gretton2012akernel}. KID represents the deviation of the distribution of the generated realizations from the distribution of the true test data, and can be conveniently computed when data is high-dimensional.
Given a non-negative and symmetric kernel function $k:\R^D\times\R^D\to\R$ and data $\{\bm x_{i}\}_{i=1}^T$ and $\{\bm y_{i}\}_{i=1}^T$, the KID is defined as
\begin{equation}
\label{eq:kid-main}
\begin{aligned}
    \KID (\{\bm x_{i}\}_{i=1}^{T},\{\bm y_{j}\}_{j=1}^{T}) 
    &= \frac{1}{T(T-1)}\sum_{\substack{i,j=1 \\ i\neq j}}^T k(\bm x_{i},\bm x_{j}) - \frac{2}{T^2}\sum_{i=1}^T\sum_{j=1}^T k(\bm x_{i},\bm y_{j})\\
    &\quad +\frac{1}{T(T-1)}\sum_{\substack{i,j=1 \\ i\neq j}}^T k(\bm y_{i},\bm y_{j}).
\end{aligned}
\end{equation}
Following \cite{binkowski2018demystifying}, the kernel function we choose is the rational quadratic kernel
\begin{align}
    k_{\text{rq}}(\bm x_{i},\bm y_{j}) \coloneqq \sum_{\ell\in\mathcal{I}}\bigg(1+\frac{\lVert \bm x_{i} - \bm y_{j}\rVert^2}{2\ell}\bigg)^{-\ell},
\end{align}
where $\mathcal{I}=\{0.2,0.5,1.0,2.0,5.0\}$ is a mixture of length scales to balance the bias effects from the different values. 
In order to evaluate the efficacy of the BF-VAE and HF-VAE models, we generated new realizations from the trained BF-VAE and HF-VAE models, denoted by $\{\bm x_{i}^\BF\}_{i=1}^{T},\{\bm x_{j}^\HF\}_{j=1}^{T}$, respectively, where $T$ is the test data size. 
The KIDs between the generated realizations and the true test data ${\{\bm x^H_l\}}_{l=1}^{T}$ are computed as
\begin{align}
    \KID^\BF &\coloneqq \KID(\{\bm x_l^H\}_{l=1}^{T},\{\bm x_i^\BF\}_{i=1}^{T}),\\
    \KID^\HF &\coloneqq \KID(\{\bm x_l^H\}_{l=1}^{T},\{\bm x_j^\HF\}_{j=1}^{T}).
\end{align}
Further discussion and technical details regarding KID are presented in \ref{apdx: mmd}. 
In addition to KID, we provide 1,000 synthesized  QoI realizations generated by both the trained HF-VAE and BF-VAE models, along with their corresponding true HF counterparts. This additional result serves to further validate the KID outcomes. The hyperparameter $\beta$ in Equation~\eqref{eq:bf-vib-loss} is tuned to reduce the KID value of the LF-VAE and, as discussed in Section~\ref{ssec:hyper-tune}, $\gamma$ in Equation~\eqref{eq:repara3} is assumed to be zero. 

\subsection{Composite Beam}
\label{ssec:beam}
\begin{figure}[!ht]
	\centering
	\includegraphics[trim = 33mm 80mm 28mm 60mm, clip, width = 1.0\textwidth]{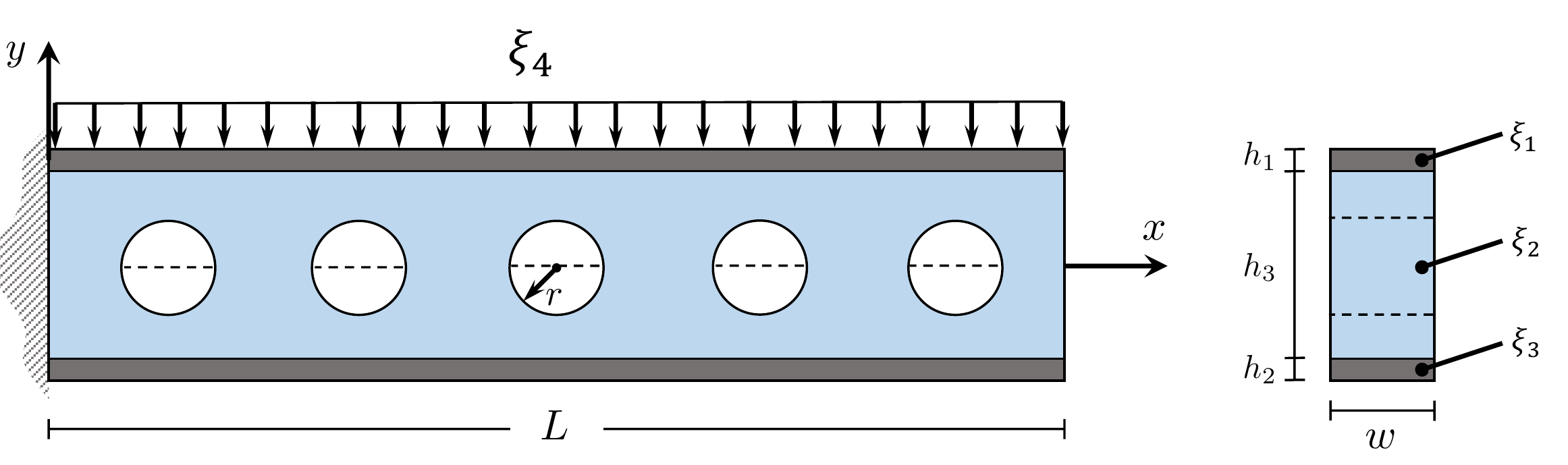}
	\caption{Cantilever beam (left) and the composite cross section (right) adapted from \cite{hampton2018practical}. }
	\label{fig:cantilever}
\end{figure}

Following \cite{hampton2018practical,de2020transfer,de2022neural,cheng2022quadrature}, we consider a plane stress, cantilever beam with composite cross section and hollow web, as shown in Figure \ref{fig:cantilever}. 
The quantities of interest, in this case, are the displacements of the top cord at 128 equi-spaced points and represented as a vector with 128 entries.  
The uncertain inputs of the model are denoted as $\bm\xi=(\xi_1, \xi_2, \xi_3, \xi_4)$, where $\xi_1$, $\xi_2$ and $\xi_3$ are the Young's moduli of the three components of the cross section and $\xi_4$ is the intensity of the applied distributed force on the beam; see Figure~\ref{fig:cantilever}. These are assumed to be statistically independent and uniformly distributed. 
The range of the input parameters, as well as the other deterministic parameters, are provided in Table~\ref{table:beam}.

\begin{figure}[htbp]
	\centering
	\includegraphics[width = 0.8\textwidth]{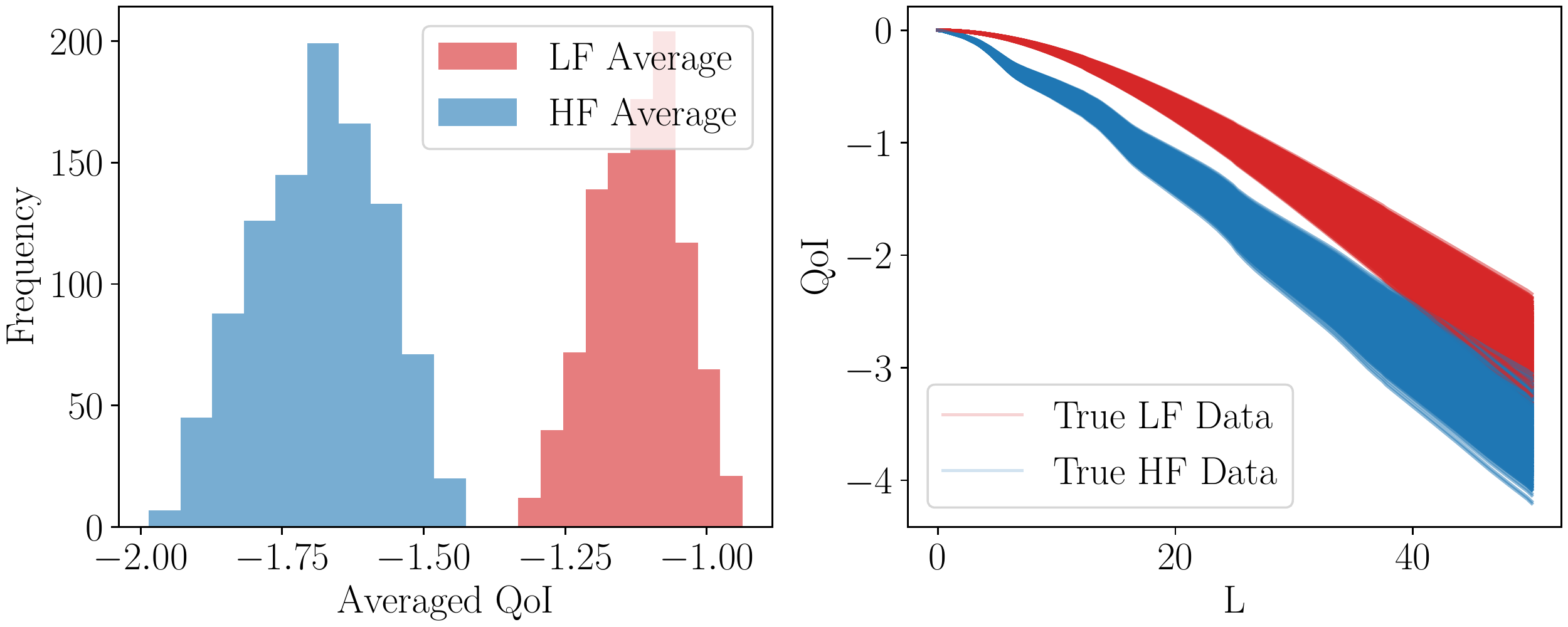}
	\caption{A histogram of the averaged QoI solutions along 128 spatial points from the LF and HF composite beam models (left) and 1,000 realizations of LF and HF QoIs (right).}
	\label{fig:beam-bf}
\end{figure}

\begin{figure}[htbp]
	\centering
	\includegraphics[trim = 0mm 80mm 0mm 70mm, clip, width = 0.8\textwidth]{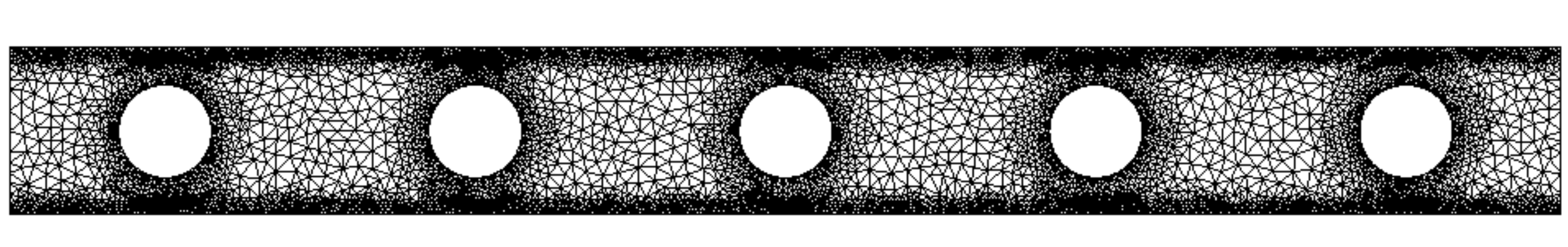}
	\caption{Finite element mesh used to generate HF solutions.  }
	\label{fig:mesh}
\end{figure}

\begin{table}[htb]
	\centering
	\caption{The values of the parameters in the composite cantilever beam model. 
		The centers of the holes are at $x=\{5,15,25,35,45\}$.
		The entries of  $\bm\xi$ are drawn independently and uniformly at random from the specified intervals. 
		\label{table:beam}} 
	\centering
	\begin{tabular}{ c c c c c c c c c c} 
		\toprule 
			$L$ & $h_1$ & $h_2$ & $h_3$ & $w$ & $r$ & $\xi_1$ & $\xi_2$ & $\xi_3$ & $\xi_4$\\
		\midrule
			50  &  0.1 & 0.1 & 5 &  1 &  1.5 & $[0.9\text{e6}, 1.1\text{e6}]$  &  $[0.9\text{e6}, 1.1\text{e6}]$ & $[0.9\text{e4}, 1.1\text{e4}]$  & $[9,11]$\\ 
		\bottomrule
	\end{tabular} 
	\label{table:parameters} 
\end{table}
The HF QoI $\bm x^H$ is based on a finite element discretization of the beam using a triangular mesh, as Figure~\ref{fig:mesh} shows. 
The LF QoI $\bm x^L$ is derived from the Euler--Bernoulli beam theory in which the vertical cross sections are assumed to remain planes throughout the deformation. 
The LF model ignores the shear deformation of the web and does not take the circular holes into account, which makes the LF results smoother than their HF counterparts, as displayed in Figure~\ref{fig:beam-bf}. 
Considering the Euler--Bernoulli theorem, the vertical displacement $u$ is 
\begin{equation}
	\label{eq:Euler-Bernoulli}
	EI_n\frac{d^4u(x)}{dx^4}=-\xi_4,
\end{equation}
where $E$ and $I_n$ are, respectively, the Young's modulus and the moment of inertia of an equivalent cross section consisting of a single material. We let $E=\xi_3$, and the width of the top and bottom sections are $w_1=(\xi_1/\xi_3)w$ and $w_2=(\xi_2/\xi_3)w$, while all other dimensions are the same, as Figure \ref{fig:cantilever} shows. The solution of \eqref{eq:Euler-Bernoulli} is 
\begin{equation}
\label{eq:beam-lf}
	u(x)=-\frac{qL^4}{24EI_n}\left(\left(\frac{x}{L}\right)^4-4\left(\frac{x}{L}\right)^3+6\left(\frac{x}{L}\right)^2\right).
\end{equation}
Since the LF data are directly obtained through an explicit formula in Equation~\eqref{eq:beam-lf}, its computational cost is negligible. 

The VAE models are implemented using fully-connected neural networks for both the encoder and decoder, each with two hidden layers and widths of 64 and 16 units, and GeLU activation functions. 
The latent space dimension is fixed at 4. 
The optimization of the VAE models is performed using the Adam optimizer with a learning rate of $1\times 10^{-3}$ and Adam-betas of $0.9$ and $0.99$. 
The batch size is set to 64, and the number of epochs for the initial training of the LF-VAE (line \ref{algline:lf-trn} in Algorithm~\ref{alg:BFVAE}) is 2,000, followed by 1,000 for the BF-VAEs (line \ref{algline:bf-trn} in Algorithm~\ref{alg:BFVAE}). 
The hyperparameter $\beta$ is $0.04$.

An LF-VAE model is first trained with $N = \text{4,000}$ samples drawn from $p(\bm x^L)$. 
Since LF data are directly generated from Equation~\eqref{eq:beam-lf}, the cost of LF data is trivial and can be ignored. 
A BF-VAE is built with parameters $\bm\phi$ and $\bm\theta$ initialized from the trained LF-VAE following Algorithm~\ref{alg:BFVAE}.
We examine the performance of BF-VAE as a function of the number of HF training samples, with HF-VAE trained solely on the same HF data as a baseline.
The KID performance is evaluated using 1,000 test data and 1,000 samples from each of the trained VAEs across 10 trials, with the results averaged over the trials.

\begin{figure}[htbp]
	\centering
	\includegraphics[width = 0.8\textwidth]{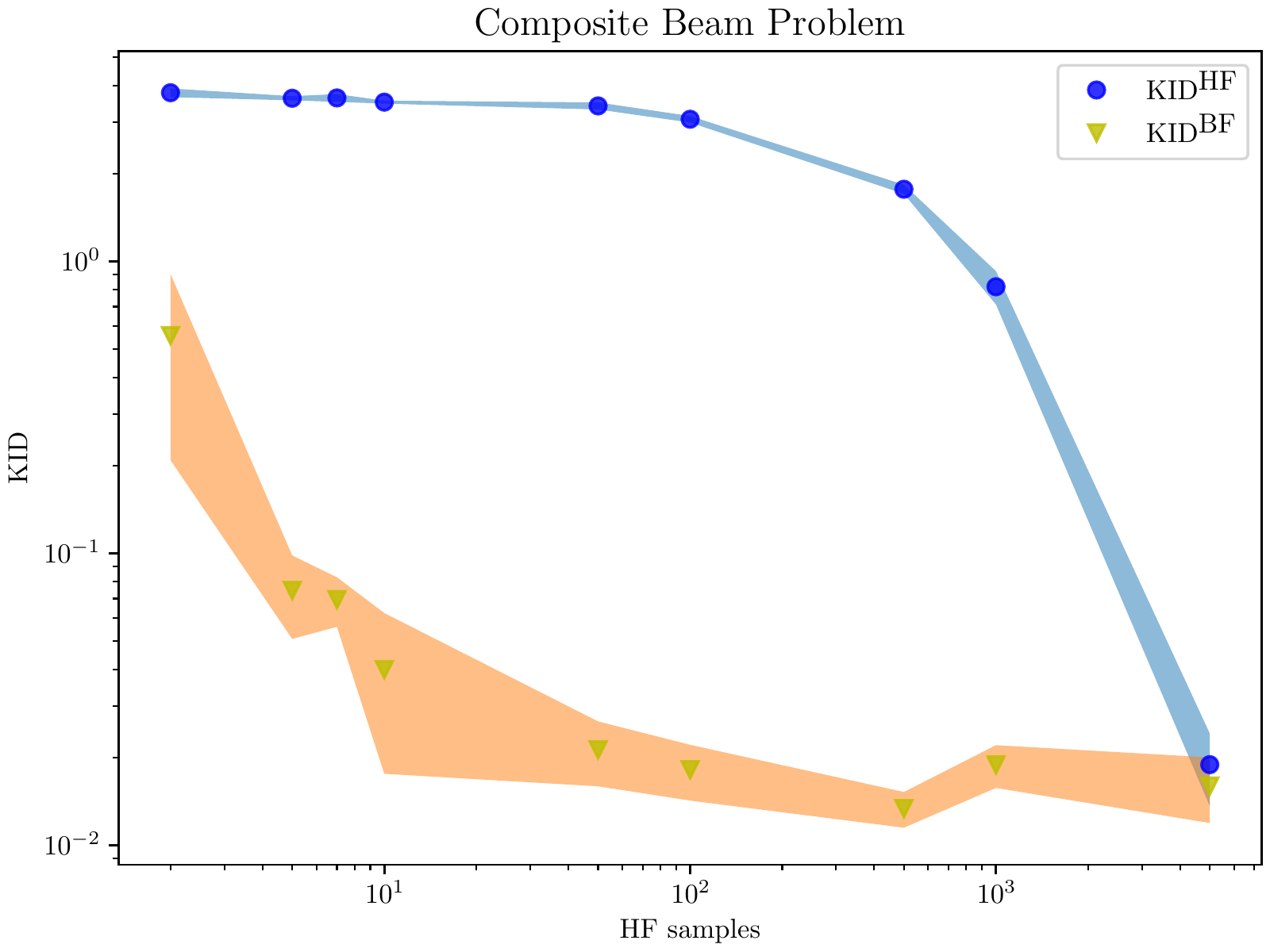}
	\caption{The KID results for the composite beam example given different sizes of HF data. 
    Each circle represents the average KID between test data and the VAEs' realizations over 10 separate trials. 
    The shaded area is half the empirical standard deviation of these 10 trials.}
	\label{fig:beam-mmd}
\end{figure}

\begin{figure}[htbp]
	\centering
	\includegraphics[trim = 30mm 0mm 30mm 0mm, clip, width = 1.0\textwidth]{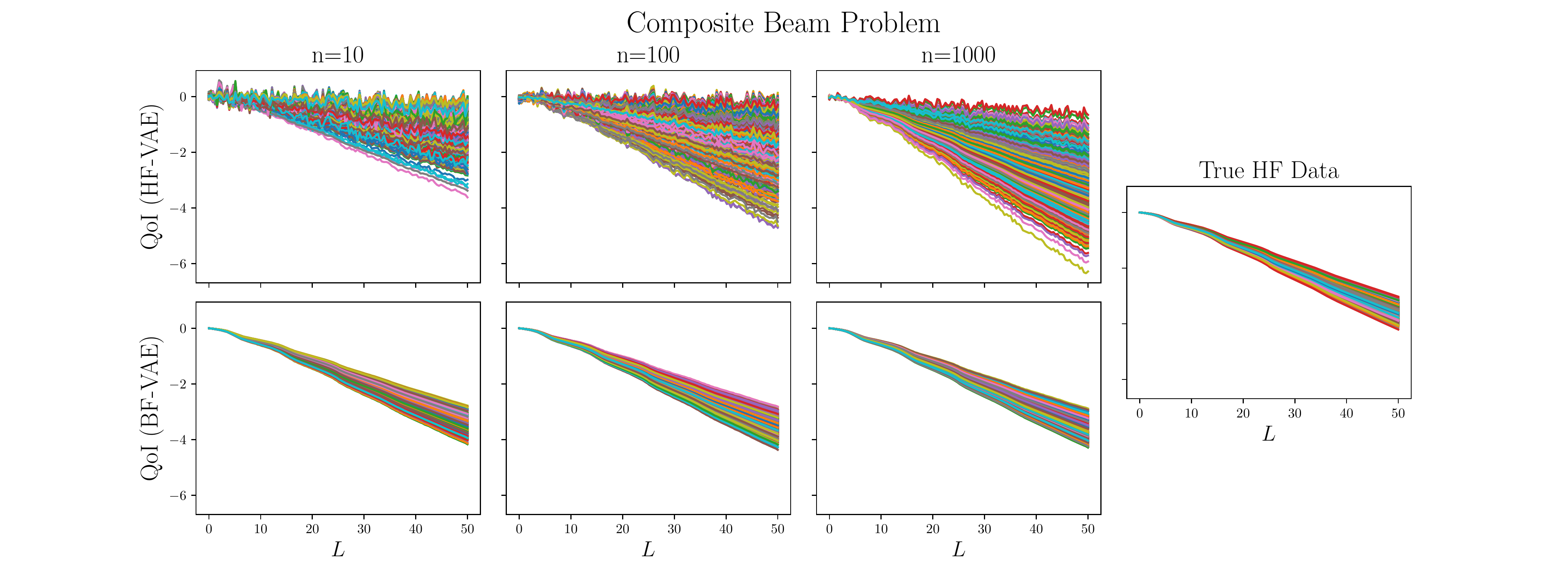}
	\caption{Comparison of 1,000 samples generated from the trained HF-VAE (top row), BF-VAE (bottom row) and the true HF model (right).
    A different number of HF realizations are used in each of the first three columns: $n=10$ (left column), $n=100$ (middle left column), and $n=\text{1,000}$ (middle right column).}
	\label{fig:beam-com}
\end{figure}

Figure~\ref{fig:beam-mmd} illustrates the KID performance of both BF-VAE and HF-VAE, with the x-axis representing the number of HF data used for training and y-axis being KID values evaluated following Equation~\eqref{eq:kid-main}. 
The results show that $\KID^\BF$ begins to converge with a small number of HF data, while $\KID^\HF$ only starts to converge when the number of HF data exceeds 100. 
Given the practical limitations on the acquisition of HF data, the superiority of the BF-VAE model is thus evident. 
Figure~\ref{fig:beam-com} presents 1,000 realizations drawn from the trained HF-VAE and BF-VAE. 
We expect the displacement as a function of horizontal distance to be smooth, but the HF-VAE samples fail to present these properties when $n<1,000$ and thus, unphysical. 
It shows that BF-VAE is able to provide a reliable result with only a small number of HF training samples, while HF-VAE requires more HF data to converge. 
Both figures demonstrate the effectiveness of the BF-VAE algorithm in utilizing the information from LF data to 
estimate the distribution of the HF QoI.

%
\subsection{Cavity Flow}
\label{ssec:cavity}

\begin{figure}[htbp]
    \centering
    \includegraphics[width=0.4\textwidth]{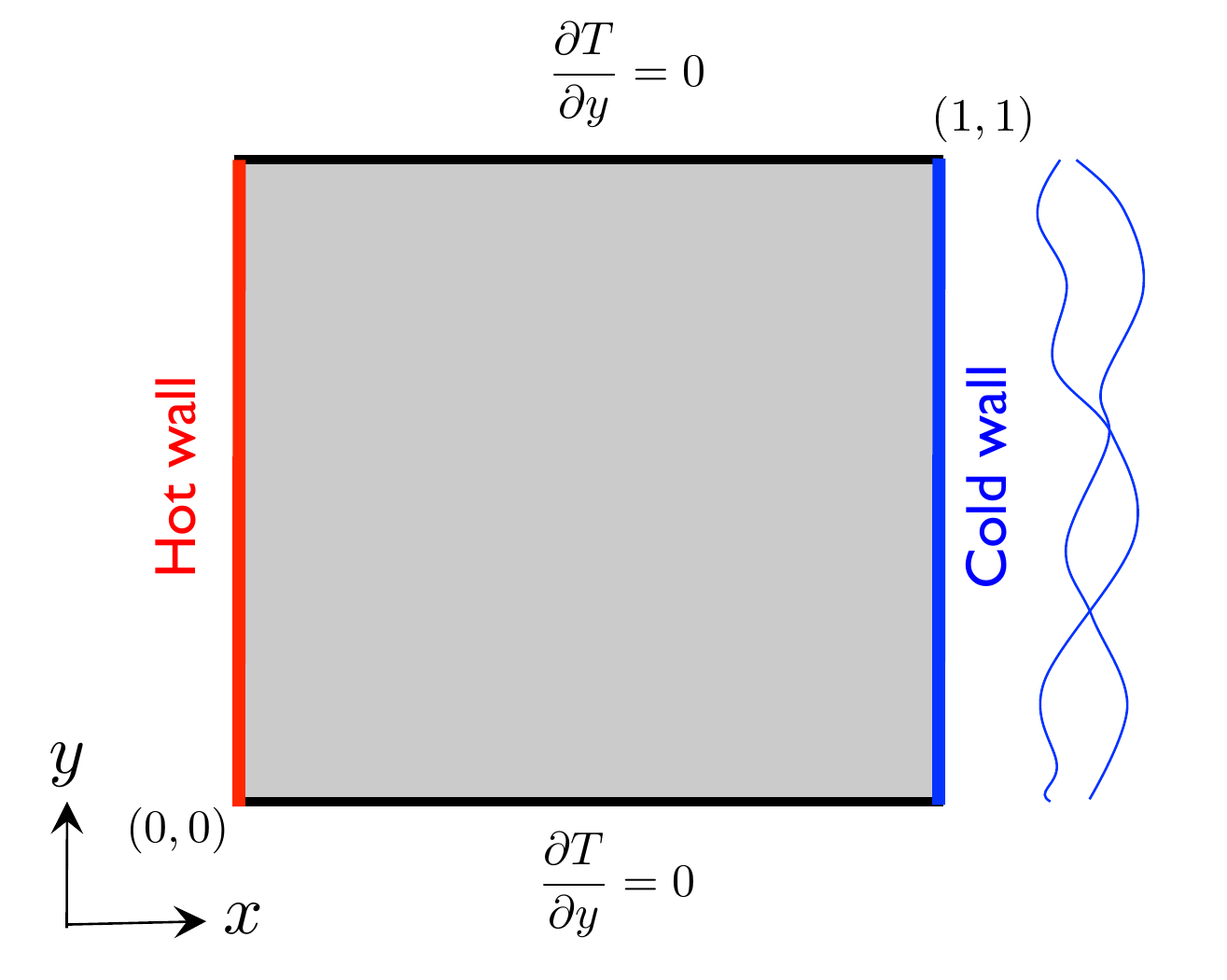}
    \caption{A figure of the temperature-driven cavity flow problem, reproduced from Figure~5 of \cite{fdki17}.}
    \label{fig:cavity_scheme_color}
\end{figure}

Here we consider the case of the temperature-driven fluid flow in a 2D cavity, with the quantity of interest being the heat flux along the hot wall as Figure~\ref{fig:cavity_scheme_color} shows. 
The left-hand wall is considered as the hot wall with a random temperature $T_h$, while the right-hand wall, referred to as the cold wall, has a smaller random temperature $T_c$ with a constant mean of $\bar{T}_c$. 
The horizontal walls are treated as adiabatic. 
The reference temperature and the temperature difference are given by $T_{\textup{ref}}=(T_h+ \bar{T}_c)/2$ and $\Updelta T_{\textup{ref}}=T_h-\bar{T}_c$, respectively. 
The normalized governing equations are given by
\begin{equation}\label{eqn:cavity}
\begin{aligned}
	&\frac{\partial \u}{\partial t} + \u\cdot\nabla \u=-\nabla p +\frac{\text{Pr}}{\sqrt{\text{Ra}}}\nabla^2 \u+\text{Pr}\Theta\textbf{e}_y,\\
	&\nabla\cdot\u=0,\\
	&\frac{\partial \Theta}{\partial t}+\nabla\cdot(\u\Theta)=\frac{1}{\sqrt{\text{Ra}}}\nabla^2\Theta,
\end{aligned}
\end{equation}
 where $\textbf{e}_y$ is the unit vector $(0,1)$, $\u=(u,v)$ is the velocity vector field, $\Theta=(T-T_{\textup{ref}})/\Updelta T_{\textup{ref}}$ is normalized temperature, $p$ is pressure, and $t$ is time. 
 The walls are subject to no-slip boundary conditions.
 The dimensionless Prandtl and Rayleigh numbers are defined as $\text{Pr}=\nu_\text{visc}/\alpha$ and $\text{Ra}=g\tau\Updelta T_{\textup{ref}}W^3/(\nu_\text{visc}\alpha)$, respectively, 
where $W$ is the width of the cavity, $g$ is gravitational acceleration, $\nu_\text{visc}$ is kinematic viscosity, $\alpha$ is thermal diffusivity, and $\tau$ is the coefficient of thermal expansion. We set $g=10$, $W=1$, $\tau=0.5$, $\Updelta T_{\textup{ref}}=100$, $\text{Ra}=10^6$, and $\text{Pr}=0.71$. On the cold wall, we apply a temperature distribution with stochastic fluctuations as 
\begin{equation}
    T(x=1,y)=\bar{T}_c+\sigma_T\sum_{i=1}^{M} \sqrt{\lambda_i}\varphi_i(y)\xi_i,
\end{equation}
where $\bar{T}_c=100$ is a constant, 
$\{\lambda_i\}_{i\in[M]}$ and $\{\varphi_i(y)\}_{i\in[M]}$ are the $M$ largest eigenvalues and corresponding eigenfunctions of the kernel $k(y_1,y_2)=\exp(-|y_1-y_2|/0.15)$, and each $\xi_i \overset{\text{i.i.d.}}{\sim} U[-1, 1]$.
We let the input dimension $M=52$ and $\sigma_T=2$. 
The vector $\bs{\xi} = (\xi_1,\dots , \xi_{52})$ is the uncertain input of the model. 
These considerations align with previous works in \cite{bc15,peng14weighted,hd15,hd15_2,hd18,cheng2022quadrature}. 

\begin{figure}[htbp]
	\centering
	\includegraphics[width = 0.8\textwidth]{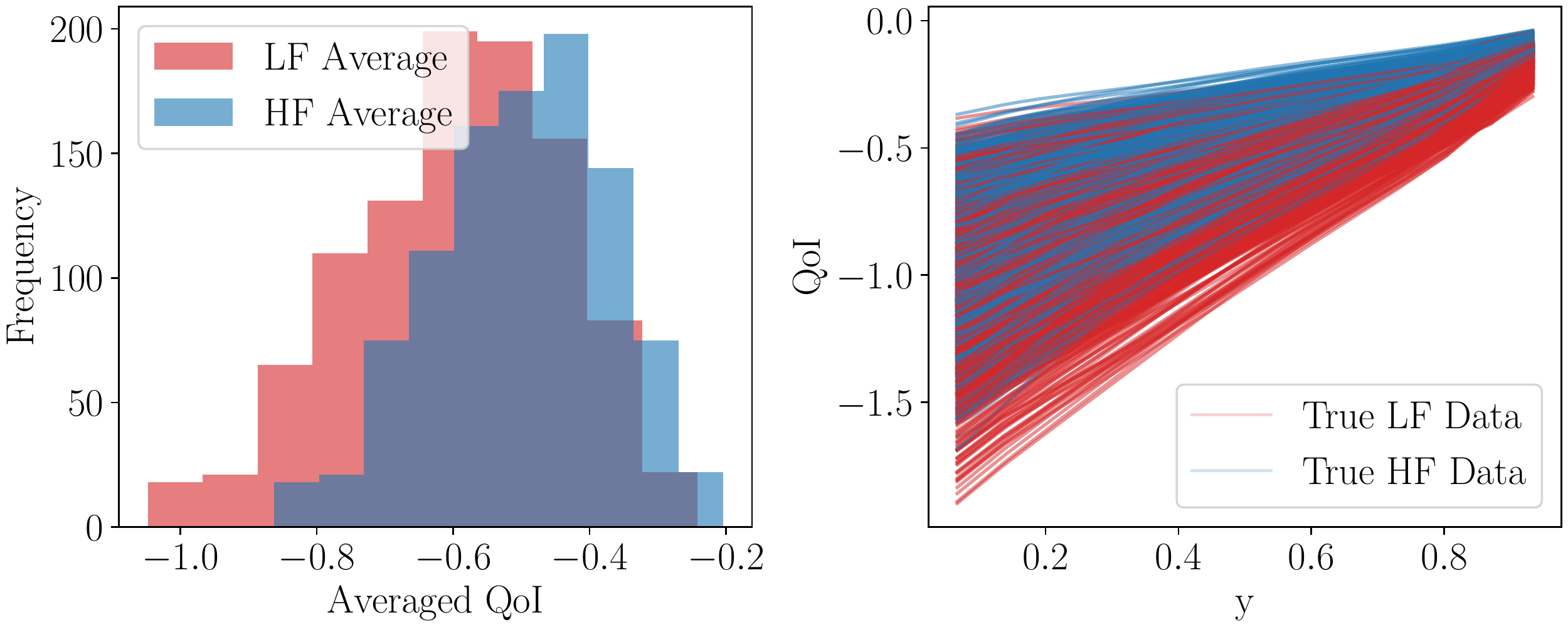}
	\caption{A histogram of the QoI solutions averaged on all spatial points from the LF and HF cavity flow models is shown in the left figure and 1,000 LF and HF QoIs are presented in the right figure.}
	\label{fig:cav-bf}
\end{figure}

Unlike the composite beam problem, the low-fidelity model is based on a coarser spatial discretization of the governing equation. Specifically, we employ the finite volume method with a grid of size $256 \times 256$ to produce the HF QoI $\bm x^H$ and a coarser grid of size $16 \times 16$ to produce the LF QoI $\bm x^L$. 
A comparison of LF and HF estimates of the QoI are presented in Figure~\ref{fig:cav-bf}. 
Based on the analysis from \cite{fdki17}, the HF/LF ratio of the computational cost for this problem is $9410.14$, which means the time for computing one HF realization is equivalent to the time for computing approximately $9410$ LF realizations. 
Since the auto-encoder structure requires both LF and HF input data to have the same dimension, we interpolate the LF data linearly on the fine grid and let the QoI be the (interpolated) steady-state heat flux along the hot wall at 221 equispaced points over $[0.067,0.933]$, including the endpoints. 
For the VAE models, we use fully connected neural networks to model the encoder and decoder with ReLU activation functions, three hidden layers, and internal widths 221--128--64--16 determined by some preliminary tests. 
The dimension of the latent space is 4. 
The number of LF samples used for training the LF-VAE is $N=\text{4,000}$.
The cost of generating these 4,000 samples is equal to 42\% of the cost of generating a single HF realization. 
As this equivalent cost is sufficiently small compared to the number of HF realizations we used for testing, we ignore the cost of generating the LF data. 
The optimizer is Adam with a learning rate $1\times 10^{-3}$ and Adam-betas $0.9,0.99$. 
The batch size for the optimization is set to 64. 
The epoch number is 2,000 for the initial LF-VAE training (line \ref{algline:lf-trn} in Algorithm~\ref{alg:BFVAE}) followed by 1,000 epochs for BF-VAE training (line \ref{algline:bf-trn} in Algorithm~\ref{alg:BFVAE}). 
The value of the hyperparameter $\beta$ is set to 4.5. 

\begin{figure}[htbp]
	\centering
	\includegraphics[width = 0.8\textwidth]{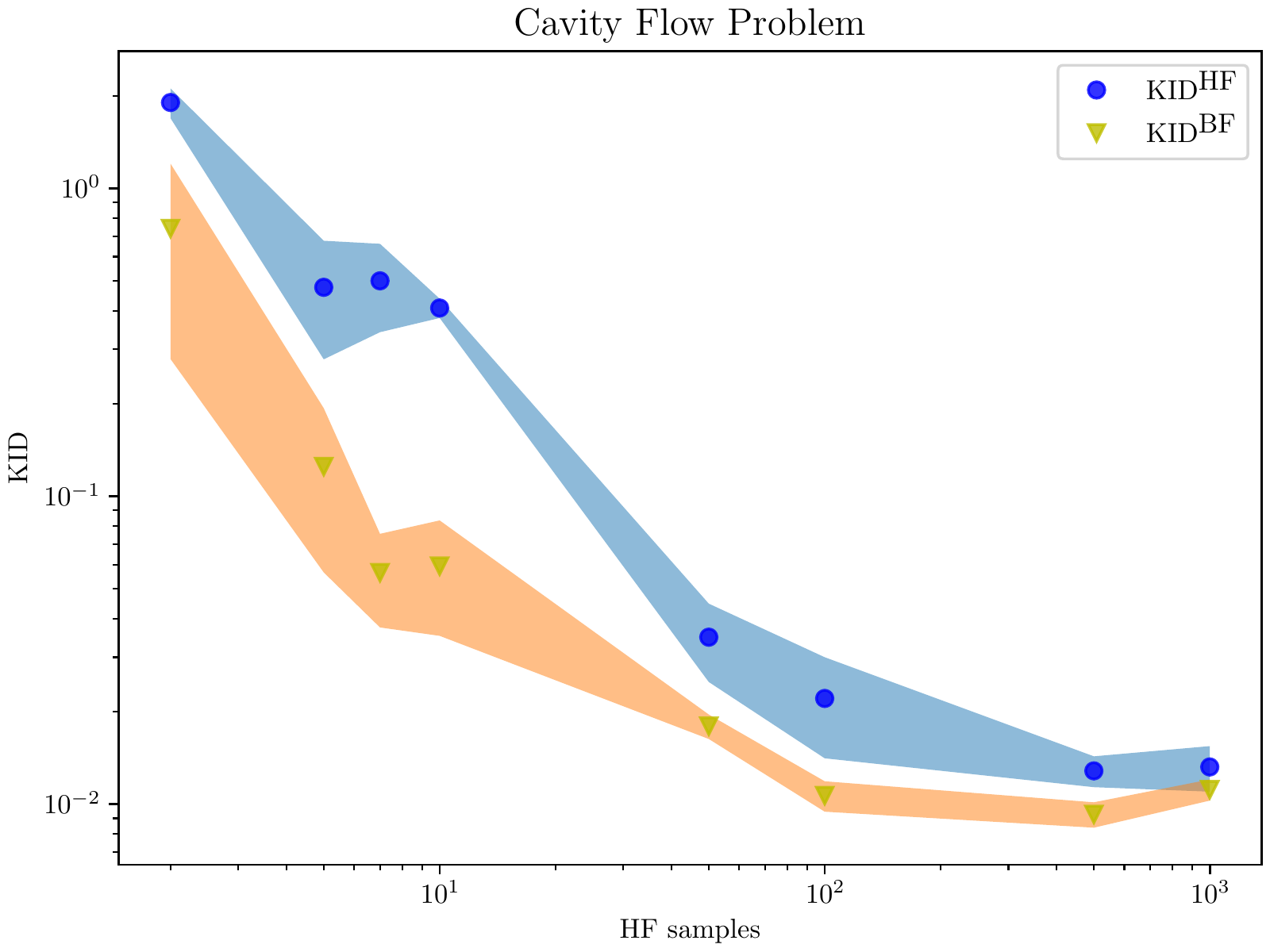}
	\caption{The KID result for the cavity flow problem given different sizes of HF data. Each point represents the average KID between test data and the VAEs' realizations over 10 separate trials. The shaded area corresponds to half the empirical standard deviation of these 10 trials.}
	\label{fig:cav-mmd}
\end{figure}

\begin{figure}[htbp]
	\centering
	\includegraphics[trim = 30mm 0mm 30mm 0mm, clip, width = 1.0\textwidth]{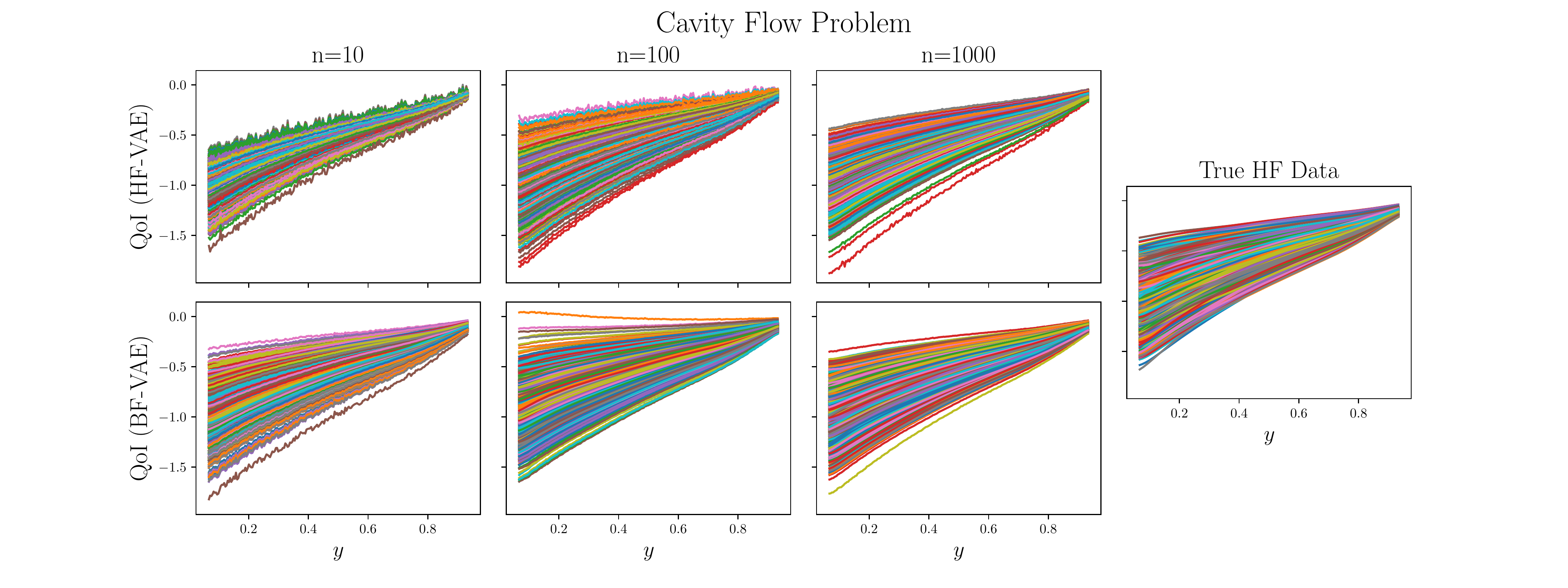}
	\caption{Comparison of 1,000 samples generated from the trained HF-VAE (top row), BF-VAE (bottom row) and the true HF model (right).
    A different number of HF realizations are used in each of the first three columns: $n=10$ (left column), $n=100$ (middle left column), and $n=\text{1,000}$ (middle right column).}
	\label{fig:cav-com}
\end{figure}

The average KID between HF data and data generated by the HF-VAE and BF-VAE, for different numbers of HF training samples sizes, are shown in Figure~\ref{fig:cav-mmd}. 
The averages are computed over 10 trials between 1,000 real samples and 1,000 VAE-simulated realizations.  
Newly generated realizations of HF-VAE and BF-VAE based on different HF training sample sizes are shown in Figure~\ref{fig:cav-com}. 
The result of Figure~\ref{fig:cav-mmd} indicates that $\KID^\BF$ is consistently lower than $\KID^\HF$ but gets closer when more HF data is available. 
Figure~\ref{fig:cav-com} suggests that BF-VAE returns smoother and more reliable predictions compared to those of HF-VAEs, especially with limited HF training data, which means BF-VAE produces more realistic results.
Both figures reveal that BF-VAE has better performance than HF-VAE when the two models are given the same amount of HF training data. 


\subsection{Burgers' Equation}
\label{ssec:burgers}

\begin{figure}[htbp]
	\centering
	\includegraphics[width = 0.8\textwidth]{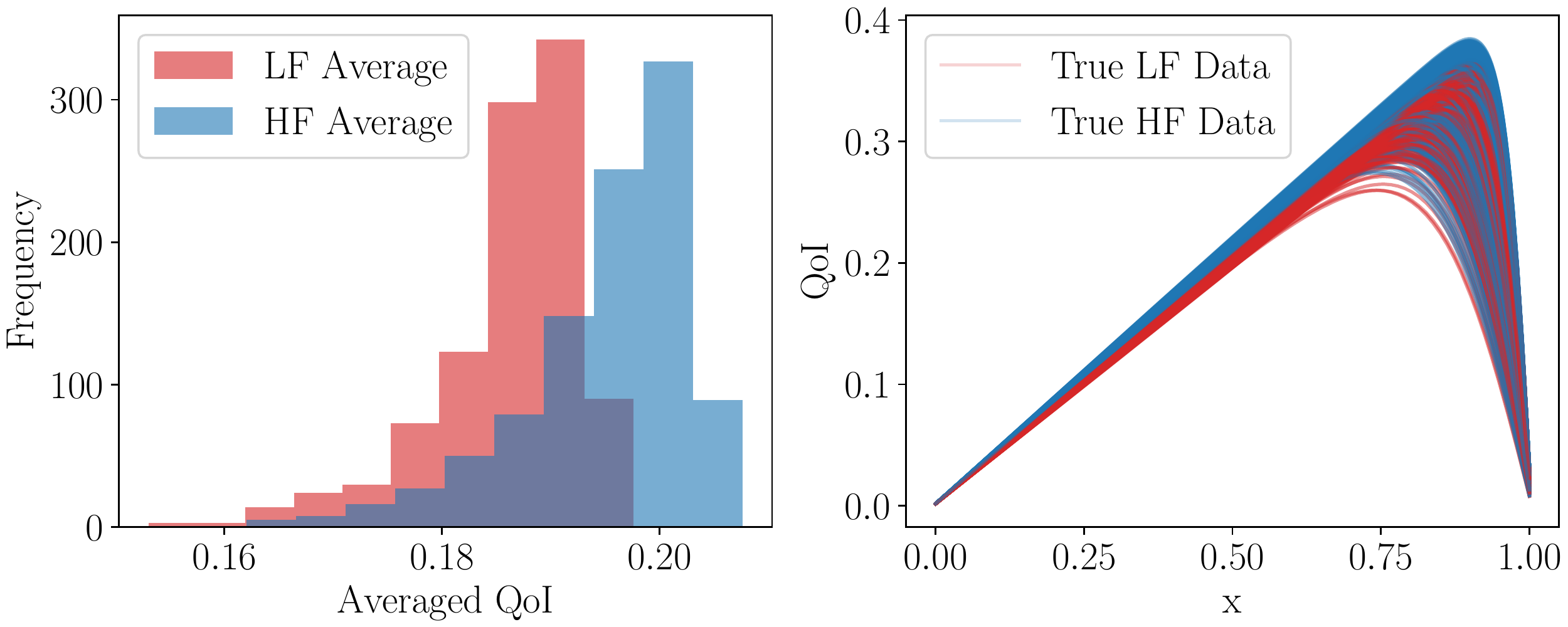}
	\caption{Histogram of the QoI values averaged on all spatial points from the LF and HF viscous Burgers' models is shown in the left figure and 1,000 LF and HF QoIs are plotted in the right figure.}
	\label{fig:burgers-bf}
\end{figure}

The last example is a one-dimensional unsteady viscous Burgers' equation with uncertain initial condition and viscosity. The random velocity field $u(x,t,\bm \xi)$ with parameters $\bm\xi$ is governed by

\begin{equation}
\label{eq:burgers-governing}
\begin{aligned}
    &\frac{\partial u(x,t,\bm \xi)}{\partial t} + u(x,t,\bm \xi)\frac{\partial u(x,t,\bm \xi)}{\partial x} = \frac{\partial}{\partial x}\bigg(\nu\frac{\partial u(x,t,\bm \xi)}{\partial x}\bigg), \quad (x,t)\in [0,1]\times[0,2], \\
    &u(0,t,\bm \xi)=u(1,t,\bm \xi)=0, \\
    &u(x,0,\bm \xi) = g(x,\bm \xi), \quad x\in[0,1],
\end{aligned}
\end{equation}
where the viscosity $\nu$ is modeled by a shifted beta random variable Beta$(0.5,5)$ over $[0.01,0.05]$. The initial condition $g(x,\bm \xi)$ is a stochastic field given by
\begin{align}
    g(x,\bm \xi) &= \sin(\pi x) + \sigma_g\sum_{k=2}^{M}\frac{1}{k}\sin(\pi kx)\xi_{k-1},\label{eq:burgers-g}
\end{align}
where $\sigma_g=1.2840\times 10^{-1}$ and $M=6$. 
The random inputs $\xi_1,\xi_2,\dots,\xi_{M-1}$ are i.i.d.\ uniformly distributed between -1 and 1, resulting in a random input vector $\bm\xi = (\xi_1,\xi_2,\dots,\xi_{M-1},\nu)$.
The QoIs are the values of $u(x,t=2,\bm \xi)$ at 254 equi-spaced $x$ nodes between 0 and 1, excluding the boundary points. 
To generate bi-fidelity data, the discretization of the Equation~\eqref{eq:burgers-governing} is carried out using two space/time grid sizes. 
The LF data is obtained using the semi-implicit, two-step Adam-Bashforth solver with a spatial grid of size $\Delta x=1.176\times 10^{-2}$ and time step size of $\Delta t=2\times10^{-2}$. 
The same solver is applied for HF data, but with smaller space/time grid sizes, $\Delta x=3.922\times 10^{-3}$ and $\Delta t=2\times10^{-4}$. 
The LF data is interpolated linearly on the finer grid so the dimensions of the LF and HF data are the same. 
The ratio of HF/LF computational cost is $98.07$. 
A comparison between LF and HF data is presented in Figure~\ref{fig:burgers-bf}. 

For the VAE implementations, we use fully connected neural networks to model the encoder and decoder, with four hidden layers as 254--256--128--64--16--4 with GeLU as activation functions.
The dimension of the latent space is 4. 
As before, the Adam optimizer with a learning rate $1\times 10^{-3}$ and Adam-betas $0.9,0.99$ is applied. 
The batch size for the optimization is 64. 
The epoch number is 2,000 for the initial LF-VAE training (line \ref{algline:lf-trn} in Algorithm~\ref{alg:BFVAE}) with an additional 1,000 epochs for BF-VAE training (line \ref{algline:bf-trn} in Algorithm~\ref{alg:BFVAE}). 
We use $N=\text{400}$ LF samples to train the LF-VAE, whose cost is equivalent to $4.08$ HF realizations and sufficiently small to be ignored in the following presented results.
The value of hyperparameter $\beta$ is set to $5\times 10^{-4}$. 

\begin{figure}[htbp]
	\centering
	\includegraphics[width = 0.8\textwidth]{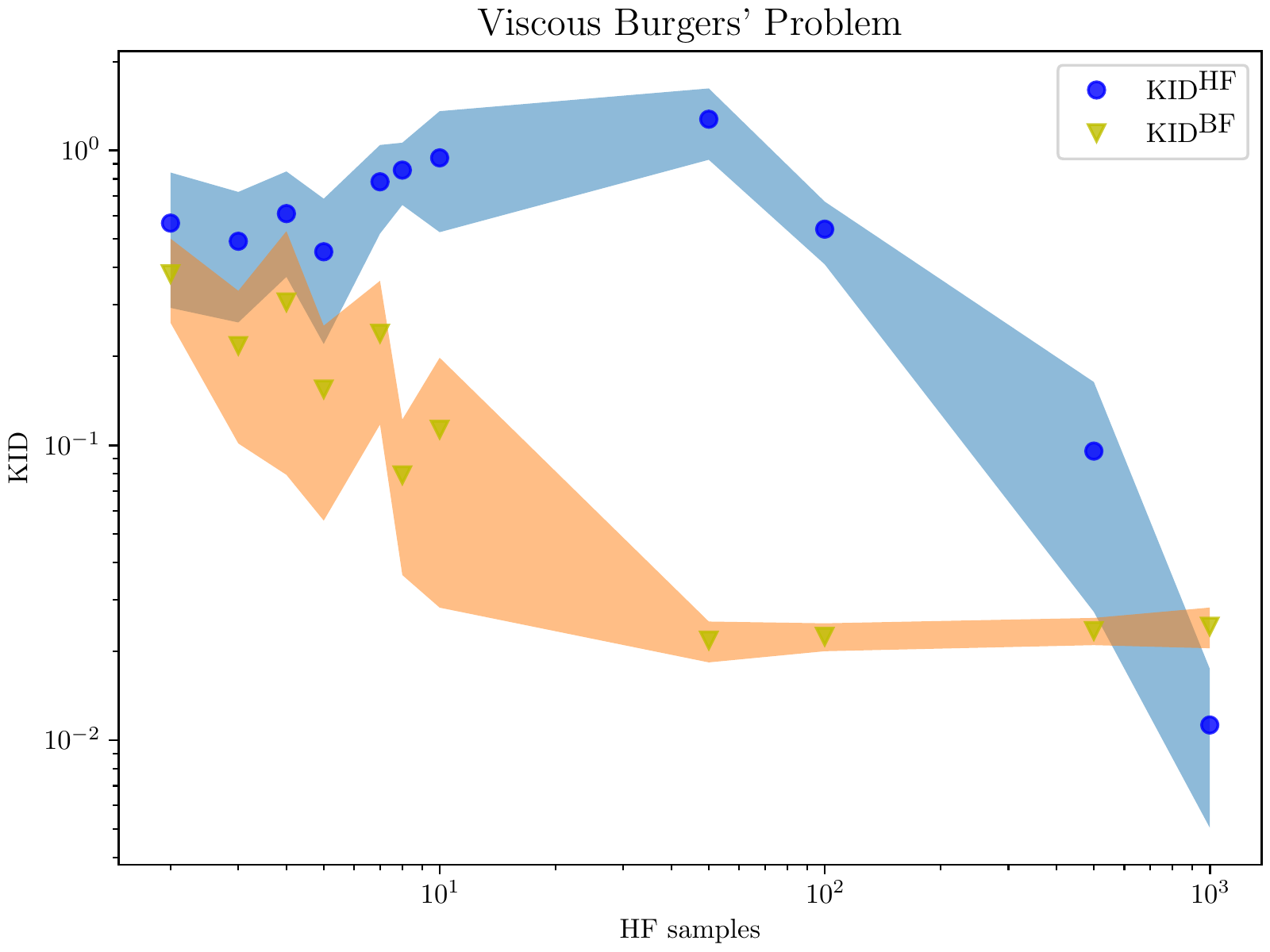}
	\caption{The $\KID$ result for the viscous Burgers' equation given different numbers of HF realizations. Each point represents the average $\KID$ between the test data and the VAEs' realizations over 10 separate trials. The shaded area corresponds to half the empirical standard deviation of these 10 trials.}
	\label{fig:burgers-mmd}
\end{figure}

\begin{figure}[htbp]
	\centering
	\includegraphics[trim = 30mm 0mm 30mm 0mm, clip, width = 1.0\textwidth]{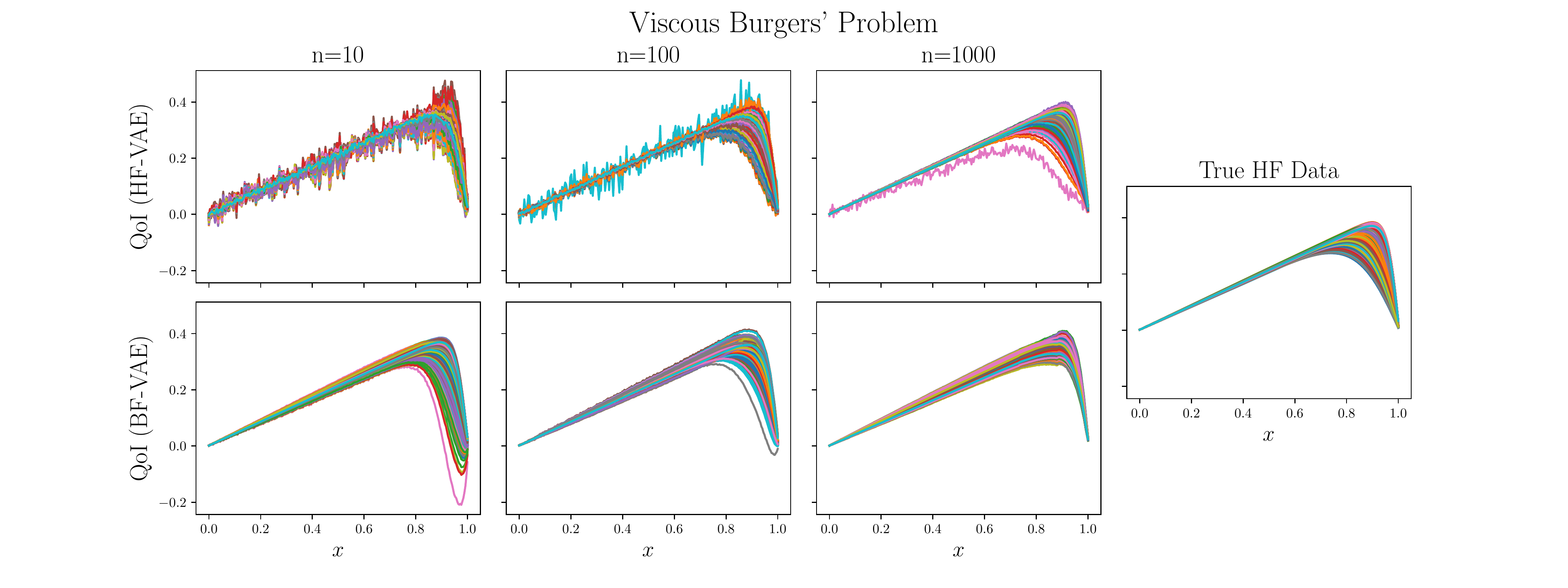}
	\caption{Comparison of 1,000 samples generated from the trained HF-VAE (top row), BF-VAE (bottom row) and the true HF model (right).
    A different number of HF realizations are used in each of the first three columns: $n=10$ (left column), $n=100$ (middle left column), and $n=1000$ (middle right column).}
	\label{fig:burgers-com}
\end{figure}

To validate the performance of the BF-VAE model, we compare its results with those of the HF-VAE model using KID. 
The KID results are computed as the average over ten trials consisting of 1,000 test samples and 1,000 VAE-generated samples, which are presented in Figure~\ref{fig:burgers-mmd}. 
Additionally, we demonstrate the validity of the BF-VAE model by generating realizations and comparing them with the HF-VAE counterparts, as shown in Figure~\ref{fig:burgers-com}. 
Based on our evaluation, we observe that the BF-VAE model achieves better accuracy in estimating the HF QoI when $n$ is small ($<100$).
We also observe that when the size of HF data is large, e.g., more than 600, $\KID^\HF$ surpasses $\KID^\BF$ and achieves a better accuracy. 
This is typical of multi-fidelity strategies and explanations are available in \cite{de2020transfer}. 

%

\section{Conclusion}
\label{sec:conclusion}
This paper presents a novel deep generative model, the bi-fidelity variational auto-encoder (BF-VAE), for generating synthetic realizations of spatio and/or temporal QoIs from parametric/stochastic PDEs through bi-fidelity data.  With an autoencoder architecture, BF-VAE exploits a low-dimensional latent space for bi-fidelity auto-regression, which significantly reduces the number of high-fidelity (HF) samples required for training. As such, the construction of the BF-VAE model is largely independent of the dimension of the stochastic input and applicable to QoIs that do not admit low-rank representations. A training criterion for BF-VAE is proposed and analyzed using information bottleneck theory \cite{tishby2000information}. 
The empirical experiments demonstrate the efficacy of the proposed algorithm in scenarios when the amount of HF data is limited.

VAE-based approaches, including BF-VAE, typically impose a multivariate Gaussian distribution on the encoder. 
As discussed in Section~\ref{ssec:error-analysis}, this results in approximation errors.
An interesting future research direction is to try using other deep generative models that do not suffer from this shortcoming in bi-fidelity UQ applications.
Examples of promising models that have achieved state-of-the-art results in other domains include normalizing flows and diffusion models.

\section*{Acknowledgments} 

N.\ Cheng and A.\ Doostan were supported by the AFOSR awards FA9550-20-1-0138 with Dr.\ Fariba Fahroo as the program manager. 
O.\ A.\ Malik was supported by DOE award DE-AC02-05CH11231.
S.\ Becker was supported by DOE award DE-SC0023346.
The views expressed in the article do not necessarily represent the views of the AFOSR or the U.S.\ Government.
\newpage 
\bibliographystyle{elsarticle-num-names}
\bibliography{references}


\newpage
\appendix
\section{Proof of Bi-fidelity ELBO}
\label{apdx:bf-elbo}
In this section, we present the detailed proof of Equation~\eqref{eq:bf-lkh}. 
We assume the Markov property for BF-VAE so that $p(\bm x^H\vert \bm z^L,\bm z^H) = p(\bm x^H\vert \bm z^H)$ and $p(\bm z^H\vert \bm z^L,\bm x^L) = p(\bm z^H\vert \bm z^L)$. 

\begin{proof}
HF log-likelihood $\log p_{\bm\theta,\bm\psi}(\bm x^H)$ can be decomposed and lower bounded as follows 
\begin{align}
    \log p_{\bm\theta,\bm\psi}(\bm x^H)     &= \mathbb{E}_{q_{\bm\phi}(\bm z_{\bm\psi}\vert\bm x^L)}[\log p_{\bm\theta,\bm\psi}(\bm x^H)]\\
                        &= \mathbb{E}_{q_{\bm\phi}(\bm z_{\bm\psi}\vert\bm x^L)}\bigg[\log(\frac{p_{\bm\theta}(\bm x^H,\bm z_{\bm\psi})}{p_{\bm\theta}(\bm z_{\bm\psi}\vert \bm x^H)})\bigg]\\
                        &= \mathbb{E}_{q_{\bm\phi}(\bm z_{\bm\psi}\vert\bm x^L)}\bigg[\log(\frac{p_{\bm\theta}(\bm x^H,\bm z_{\bm\psi})q_{\bm\phi}(\bm z_{\bm\psi}\vert\bm x^L)}{p_{\bm\theta}(\bm z_{\bm\psi}\vert \bm x^H)q_{\bm\phi}(\bm z_{\bm\psi}\vert\bm x^L)})\bigg]\\
                        &= \KL\big(q_{\bm\phi}(\bm z_{\bm\psi}\vert\bm x^L)\|p_{\bm\theta}(\bm z_{\bm\psi}\vert\bm x^H)\big) + \mathbb{E}_{q_{\bm\phi}(\bm z_{\bm\psi}\vert\bm x^L)}\bigg[\log(\frac{p_{\bm\theta}(\bm x^H,\bm z_{\bm\psi})}{q_{\bm\phi}(\bm z_{\bm\psi}\vert\bm x^L)})\bigg]\\
                        &\geq \mathbb{E}_{q_{\bm\phi}(\bm z_{\bm\psi}\vert\bm x^L)}\bigg[\log(\frac{p_{\bm\theta}(\bm x^H,\bm z_{\bm\psi})}{q_{\bm\phi}(\bm z_{\bm\psi}\vert\bm x^L)})\bigg]\\
                        &= \mathbb{E}_{q_{\bm\phi}(\bm z_{\bm\psi}\vert\bm x^L)}\bigg[\log(\frac{p_{\bm \theta}(\bm x^H\vert\bm z_{\bm\psi})p_{\bm\psi}(\bm z^H\vert\bm z^L)p(\bm z^L)}{p_{\bm\psi}(\bm z^H\vert\bm z^L)q_{\bm\phi}(\bm z^L\vert\bm x^L)})\bigg]\\
                        &= \mathbb{E}_{q_{\bm\phi}(\bm z_{\bm\psi}\vert\bm x^L)}\bigg[\log(\frac{p(\bm z^L)}{q_{\bm\phi}(\bm z^L\vert\bm x^L)}) + \log(p_{\bm\theta}(\bm x^H\vert \bm z_{\bm\psi}))\bigg]\\
                        &= -\KL\big(q_{\bm\phi}(\bm z^L\vert\bm x^L)\|p(\bm z^L)\big) + \mathbb{E}_{q_{\bm\phi}(\bm z_{\bm\psi}\vert\bm x^L)}\big[\log\big(p_{\bm\theta}(\bm x^H\vert\bm z_{\bm\psi})\big)\big]\\
                        &= \ELBO^{\BF}(\bm\phi,\bm\psi,\bm\theta).
\end{align}
The only inequality above follows from the non-negativity of KL divergence. 
The above derivation shows that the HF log-likelihood can be lower bounded by the proposed BF-ELBO in Equation~\eqref{eq:bf-elbo}, where the tightness of the bound is controlled by the approximation error mentioned in Section~\ref{ssec:error-analysis} as
$\KL\big(q_{\bm\phi}(\bm z_{\bm\psi}\vert\bm x^L)\|p_{\bm\theta}(\bm z_{\bm\psi}\vert\bm x^H)\big)$.
\end{proof}

\section{Proof of Bi-fidelity Information Bottleneck}
\label{apdx:bf-ib}

In this section, we prove that optimizing $\ELBO^{BF}(\bm\phi,\bm\psi,\bm\theta)$ in Equation~\eqref{eq:bf-elbo} is equivalent with optimizing BF-IB objective function $\IB^\BF_\beta$ in Equation~\eqref{eq:bf-vib} with $\beta=1$.
With the BF-IB graphical model $\bm z_{\bm\psi}\leftarrow \bm x^L\leftrightarrow\bm x^H$ assumed (similar with IB in \cite{murphy2023pml}), we have $q_{\bm\phi}(\bm z_{\bm\psi}\vert\bm x^L)=q_{\bm\phi}(\bm z_{\bm\psi}\vert\bm x^L,\bm x^H)$.

\begin{proof}
\begin{align}
    &\IB^\BF(\bm\phi,\bm\psi,\bm\theta)\\
                                    &= -  \I(\bm x^L,\bm z_{\bm\psi}) + \I(\bm z_{\bm\psi},\bm x^H)\\
                                    &= - \mathbb{E}_{p(\bm x^L,\bm z_{\bm\psi})}\bigg[\log\frac{q_{\bm\phi}(\bm z_{\bm\psi},\bm x^L)}{p(\bm z_{\bm\psi})p(\bm x^L)}\bigg] + \mathbb{E}_{p(\bm x^H,\bm z_{\bm\psi})}\bigg[\log\frac{p_{\bm\theta}(\bm x^H,\bm z_{\bm\psi})}{p(\bm x^H)p(\bm z_{\bm\psi})}\bigg]\\
                                    &= -\mathbb{E}_{p_{\bm\phi}(\bm x^L,\bm z_{\bm\psi})}\bigg[\log\frac{q_{\bm\phi}(\bm z_{\bm\psi}\vert\bm x^L)}{p_{\bm\psi}(\bm z_{\bm\psi})}\bigg] + \mathbb{E}_{p(\bm x^H,\bm z_{\bm\psi})}[\log p_{\bm\theta}(\bm x^H\vert \bm z_{\bm\psi})] + \mathbb{H}[\bm x^H]\\
                                    &= -\mathbb{E}_{p(\bm x^L,\bm z^H,\bm z^L)}\bigg[\log\frac{q_{\bm\phi}(\bm z^L\vert\bm x^L)p_{\bm\psi}(\bm z^H\vert\bm z^L)}{p(\bm z^L)p_{\bm\psi}(\bm z^H\vert\bm z^L)}\bigg] + \mathbb{E}_{p(\bm x^H,\bm z^H,\bm z^L)}[\log p_{\bm\theta}(\bm x^H\vert \bm z^H)] + \mathbb{H}[\bm x^H]\\
                                    &= -\mathbb{E}_{p(\bm x^L)}\big[\KL(q_{\bm\phi}(\bm z^L\vert \bm x^L)||p(\bm z^L)\big] + \mathbb{E}_{p(\bm x^H,\bm z^H,\bm z^L)}\big[\log p_{\bm\theta}(\bm x^H\vert \bm z^H)\big] +\mathbb{H}[\bm x^H]\\
                                    &= -\mathbb{E}_{p(\bm x^L)}\big[\KL(q_{\bm\phi}(\bm z^L\vert \bm x^L)||p(\bm z^L)\big] + \int p(\bm x^H,\bm z^H,\bm z^L)\big[\log p_{\bm\theta}(\bm x^H\vert \bm z^H)\big]d\bm z^L d\bm z^H d\bm x^H +\mathbb{H}[\bm x^H]\\
                                    &= -\mathbb{E}_{p(\bm x^L)}\big[\KL(q_{\bm\phi}(\bm z^L\vert \bm x^L)||p(\bm z^L)\big]\\
                                    &\quad + \int q_{\bm\phi}(\bm z^L,\bm z^H\vert\bm x^L, \bm x^H)p(\bm x^H,\bm x^L)\big[\log p_{\bm\theta}(\bm x^H\vert \bm z^H)\big]d\bm z^H d\bm x^H d\bm x^L d\bm z^L +\mathbb{H}[\bm x^H]\\
                                    &= -\mathbb{E}_{p(\bm x^L)}\big[\KL(q_{\bm\phi}(\bm z^L\vert \bm x^L)||p(\bm z^L)\big]\\
                                    &\quad + \int q_{\bm\phi}(\bm z^L\vert\bm x^L)p(\bm x^H,\bm x^L)p_{\bm\psi}(\bm z^H\vert\bm z^L)\big[\log p_{\bm\theta}(\bm x^H\vert \bm z^H)\big]d\bm z^H d\bm x^H d\bm x^L d\bm z^L +\mathbb{H}[\bm x^H]\\
                                    &= -\mathbb{E}_{p(\bm x^L)}\big[\KL(q_\phi(\bm z^L\vert \bm x^L)||p(\bm z^L)\big] + \mathbb{E}_{p(\bm x^L,\bm x^H)}\mathbb{E}_{q_{\bm\phi}(\bm z^L\vert\bm x^L)}\big[\mathbb{E}_{p_{\bm\psi}(\bm z^H\vert\bm z^L)}[\log p_{\bm\theta}(\bm x^H\vert \bm z^H)]\big]+\mathbb{H}[\bm x^H]\\
                                    &= \mathbb{E}_{p(\bm x^L,\bm x^H)}[\ELBO^{BF}(\bm\phi,\bm\psi,\bm\theta)] + \text{constant},
\end{align}
where $\mathbb{H}[\cdot]$ is the differential entropy of the input random vector.
Since $p(\bm x^H)$ is fixed, its entropy is a constant. 
\end{proof}

\section{A Brief Introduction to KID}
\label{apdx: mmd}
In this section, we briefly introduce the kernel inception score (KID). 
KID is a commonly-used metric for evaluating the performance of generative models \cite{binkowski2018demystifying}, which stems from maximum mean discrepancy (MMD). 
MMD is a type of statistical distance that falls under the umbrella of the integral probability metric (IPM). 
Given two probability distributions $p(\bm x)$ and $q(\bm x)$, the IPM is defined as
\begin{align}
    \text{IPM}_{\Fc}(p,q) \coloneqq \sup_{f\in\Fc} \E_p[f(\bm x)] - \E_q[f(\bm x)],
\end{align}
where the function class $\Fc$ controls the value range of the IPM and $\E_p$ is the expectation with respect to $p(\bm x)$.
Larger $\Fc$ brings higher accuracy to the IPM value but also increases the computing complexity. 
By Kantorovich-Rubinstein duality theorem \cite{villani2021topics}, the Wasserstein-1 distance is a type of IPM with $\Fc$ being all Lipschitz continuous functions having Lipschitz constant bounded by 1.
However, estimating the Wasserstein distance accurately in high dimensions is difficult. 
MMD assigns $\Fc$ to be all functions in a reproducing kernel Hilbert space (RKHS) $\Hc$ with norm bounded by 1, where $\Hc$ is generated from a given kernel function $k:\R^D\times\R^D\to\R$. 
The motivation for using RKHS is for its computational convenience, as the following propositions show.

\begin{prop}
\label{prop:mmd1}
MMD can be expressed in the following alternative form.
\begin{align}
    \MMD(p,q) &\coloneqq  \sup_{\lVert f\rVert_\Hc=1} \mathbb{E}_p[f(\bm x)] - \mathbb{E}_q[f(\bm x)]\\
    &= \lVert \Gc_\Hc(p) - \Gc_\Hc(q)\rVert_\Hc,
\end{align}
where $\Gc_\Hc$ is a Bochner integral defined as $\Gc_\Hc(p)\coloneqq\int_{\mathbb{R}^D}k(\bm x,\cdot)p(\bm x)d\bm x$.
\end{prop}
The proof of Proposition~\ref{prop:mmd1} is available from the Lemma 4 in \cite{gretton2012akernel}. 
Suppose we have samples $\{\bm x_i\}_{i=1}^m\sim p(\bm x)$ and $\{\tilde{\bm x}_i\}_{i=1}^n\sim q(\bm x)$, an unbiased estimator of $\MMD^2$, named \KID, is as follow by Proposition~\ref{prop:mmd1}.
\begin{align}
\label{eq:mmd-est}
    &\KID\left(\{\bm x_i\}_{i=1}^m,\{\tilde{\bm x}_i\}_{i=1}^n\right)\\
    &= \frac{1}{m(m-1)}\sum_{\substack{i,j=1 \\ i\neq j}}^m k(\bm x_i,\bm x_j) - \frac{2}{mn}\sum_{i=1}^m\sum_{j=1}^n k(\bm x_i,\tilde{\bm x}_j) +\frac{1}{n(n-1)}\sum_{\substack{i,j=1 \\ i\neq j}}^n k(\tilde{\bm x}_i,\tilde{\bm x}_j).
\end{align}
Equation~\eqref{eq:mmd-est} discloses the connection between KID and MMD. 
In Section~\ref{sec:experiments}, we use KID to present the distributional similarity between two given samples.

The authors of \cite{binkowski2018demystifying} have shown that a rational quadratic kernel with a mixture of length scales is a superior candidate for MMD, due to its low rate of tail decay. 
The rational quadratic kernel has the form
\begin{align}
    k_{\text{rq}}(\bm x_i,\bm y_j) \coloneqq \sum_{\ell\in\mathcal{I}}\bigg(1+\frac{\lVert \bm x_i - \bm y_j\rVert^2}{2\ell}\bigg)^{-\ell},
\end{align}
where $\mathcal{I}=[0.2,0.5,1.0,2.0,5.0]$ is a mixture of different length scales. 

The following results, which appear as Theorem 10 and Corollary 16 in \cite{gretton2012akernel}, show the consistency of KID. 
\begin{prop}
\label{prop:mmd3}
Assuming both input sample sets have the same size $m$, and that the kernel function satisfies $0\leq k(\bm x,\bm y)\leq K$, KID in Equation~\eqref{eq:mmd-est} satisfies 
\begin{align}
    \P[\vert\KID-\MMD^2\vert>\epsilon]&\leq2\exp(-\frac{\epsilon^2 m}{16K^2}), \\
    m^{1/2}(\KID-\MMD^2)&\xrightarrow[]{d}\mathcal{N}(0,\sigma^2_u),
\end{align}
where $\sigma^2_u$ is a value independent of $m$ and $D$. 
\end{prop}

It should be noted that the asymptotic mean square error of KID is $m^{-1}\sigma^2_u$, which is independent of dimension $D$.
It is the key reason why we choose KID over other statistical distances to test our models, considering the problems in this work have large $D$ ($D\geq 100$). 

\end{document}